\RequirePackage{fix-cm}
\documentclass[twocolumn,10pt,final,natbib]{svjour3}

\smartqed
\usepackage{times}
\usepackage{epsfig}
\usepackage{graphicx}
\usepackage{amsmath}
\usepackage{amssymb}
\usepackage{gensymb}
\usepackage{units}
\usepackage{tabularx}
\usepackage{multirow}
\usepackage{makecell}
\usepackage{booktabs}
\usepackage{adjustbox}
\usepackage{pifont}
\usepackage{dblfloatfix}
\newcommand*\rot{\rotatebox{0}}

\usepackage[pagebackref=true,breaklinks=true,bookmarks=false,colorlinks=true,linkcolor=blue,citecolor=blue]{hyperref}

\def\makeheadbox{\relax} 

\begin{document}

\title{Spatially-Adaptive Filter Units for Compact and Efficient Deep Neural Networks}

\authorrunning{D. Tabernik et al.} 

\author{Domen Tabernik$^\text{1}$ \and Matej Kristan$^\text{1}$ \and Ale\v{s} Leonardis$^\text{2,1}$
}

\institute{D. Tabernik$^\text{1}$, M. Kristan$^\text{1}$  \at
              $^\text{1}$Faculty of Computer and Information Science\\
              University of Ljubljana, Slovenia\\
              \email{domen.tabernik@fri.uni-lj.si}\\
              \email{matej.kristan@fri.uni-lj.si}
           \and
           A. Leonardis$^\text{2,1}$ \at
              $^\text{2}$School of Computer Science, \\
              University of Birmingham, UK\\
              \email{a.leonardis@cs.bham.ac.uk}
}

\date{Received: 15 February 2019 / Accepted: 11 December 2019}

\maketitle

\begin{abstract}
 
Convolutional neural networks excel in a number of computer vision tasks. One of their most crucial architectural elements is the effective receptive field size, which has to be manually set to accommodate a specific task. Standard solutions involve large kernels, down/up-sampling and dilated convolutions. These require testing a variety of dilation and down/up-sampling factors and result in non-compact networks and large number of parameters. We address this issue by proposing a new convolution filter composed of {\em displaced aggregation units} (DAU). DAUs learn spatial displacements and adapt the receptive field sizes of individual convolution filters to a given problem, thus reducing the need for hand-crafted modifications. DAUs provide a seamless substitution of convolutional filters in existing state-of-the-art architectures, which we demonstrate on Alex\-Net, Res\-Net50, Res\-Net101, Deep\-Lab and SRN-Deblur\-Net. The benefits of this design are demonstrated on a variety of computer vision tasks and datasets, such as image classification (ILSVRC 2012), semantic segmentation (PASCAL VOC 2011, Cityscape) and blind image de-blurring (GOPRO). Results show that DAUs efficiently allocate parameters resulting in up to four times more compact networks in terms of the number of parameters at similar or better performance. 

\end{abstract}

\section{Introduction}

Deep convolutional neural networks (ConvNets)~\citep{He2015a,Xie2016c,Redmon2016,He2017} have demonstrated excellent performance across a broad spectrum of computer vision tasks, such as image classification~\citep{He2015a}, semantic segmentation~\citep{Ronneberger2015,Chen2014}, image restoration, and blind image de-blurring~\citep{Tao2018}. 
Early works showed that deep features pre-trained for one task (e.g., classification~\citep{Simonyan2015}) can be applied with some success to other tasks (e.g. semantic segmentation~\citep{Shelhamer2016}). Direct application to another task, however, is sub-optimal and architectural changes are required~\citep{Chen2016a}.

\begin{figure}
\includegraphics[width=\linewidth]{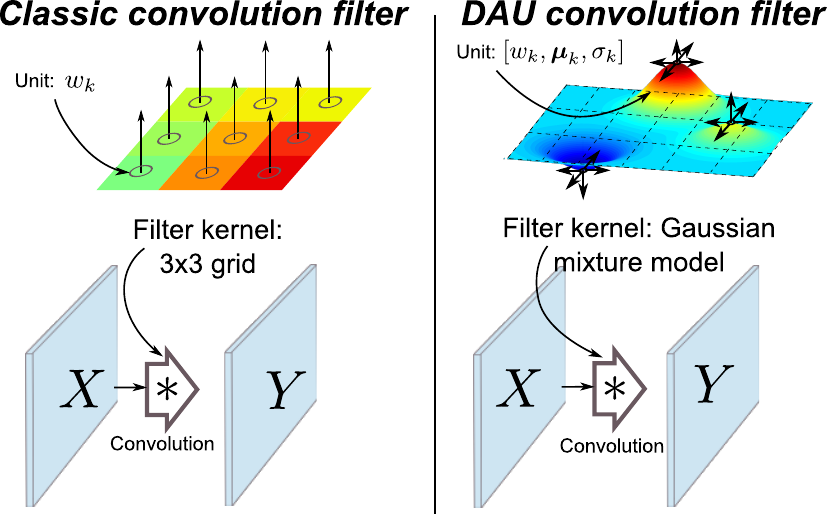}
\caption{ A classic convolution filter with a fixed-grid filter kernel (left) is replaced with a convolution filter composed of several displaced aggregation units (DAUs) whose sub-pixel positions are learned (right).
\label{fig:intro}}
\end{figure} 
 
\begin{figure*}

\includegraphics[width=\linewidth]{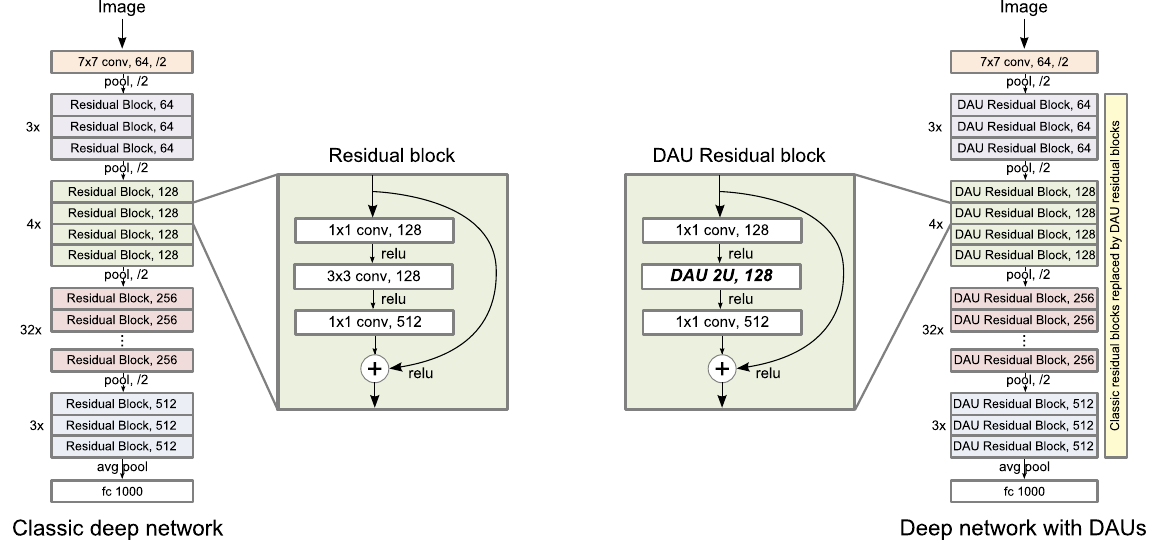}
\caption{Example of the state-of-the-art architecture (ResNet101) where classic convolution is replaced with the displaced aggregation units. In DAU residual block, DAUs with 2 units replace all $3\times3$ convolutions from the classic residual block. \label{fig:intro-resnet}}
\end{figure*} 

One of the crucial task-dependent architectural elements is the effective receptive field size of neurons~\citep{Luo2016}. In image classification, relatively small convolution filters are used and the receptive field size is increased by gradual sub-sampling of features in consecutive layers. However, the output resolution obtained by such process is too coarse for tasks that require per-pixel predictions (e.g., semantic segmentation). 

Standard strategies to address this issue are based on (i) removing the pooling layers or down-sampling, and using large convolution filters or (ii) keeping the filters small and appending a network with (gradual) up-sampling layers and skip connections~\citep{Ronneberger2015}. Both strategies significantly increase the number of parameters leading to non-compact networks and inefficient learning. \cite{Jeon2017} have proposed deforming small convolution kernels, but these deformations result in negligible change of the receptive field sizes.
 
The problem of parameter dependence on the filter size was partially addressed by~\cite{Chen2016a} with dilated convolutions. The dilation factor is manually set and fixed, which may be sub-optimal for a given application. \cite{Chen2016a} thus proposed atrous spatial pyramid pooling (ASPP), which is composed of parallel processing paths, each path with a different dilation factor. Since ASPP entails a non-negligible increase of parameters, the authors propose adding it only as a post-processing block on top of deep features. There are several drawbacks of using fixed dilations. First, the pattern of the dilation is a regular expansion of a grid-based filter. The pattern is fixed prior to learning and cannot change. Other patterns might be more appropriate for a given task, but the search would lead to a combinatorial explosion of possible filters to be tested during learning, which is practically infeasible. Secondly, large dilations significantly violate the Nyquist theorem~\citep{Amidror2013}, resulting in gridding artifacts as demonstrated by \cite{Yu2017}.

In this paper, the above issues with the effective receptive field size and filter pattern are addressed by introducing a novel formulation of the convolution filter. In particular, the standard CNN convolution filter, which is a regular grid of values, is replaced by a continuous parametric function. Specifically, a mixture of weighted Gaussian kernels is proposed for the functional form (see Figure~\ref{fig:intro}). Each Gaussian acts as a unit that aggregates feature responses locally at its displacement. During learning the displacement of units is optimized along with other parameters---hence we call these {\em displaced aggregation units} (DAU). The number of parameters in the convolution filter is thus fixed by the number of units and does not increase with the receptive field size.

Our main contribution is a DAU convolution filter (Figure~\ref{fig:intro}), which incorporates three novel concepts into the deep networks: (a) {\em decoupling of the parameter count} from the receptive field size, (b) {\em learning of the receptive field of each convolution filter} in the network, (c) and {\em automatic adjustment of the spatial focus} on the sub-feature from a previous layer through explicit modeling of the unit's position. By following those concepts the DAU-ConvNets exert compactness in terms of the number of parameters, efficient use of parameters, and allow adaptation to specific tasks without manually testing various complex dilation factors and filter patterns. Those properties directly contribute to the improved performance for various computer vision tasks.

The benefits of DAUs are demonstrated on a range of computer vision tasks (i.e., image classification, semantic image segmentation and blind image de-blurring) by replacing the standard convolution filters (see Figure~\ref{fig:intro-resnet}). We empirically verify that the resulting novel deep models:
\begin{itemize}
    \item enable automatic and {\em efficient allocation of parameters} for spatial attention while requiring as few as $25\%$ of parameters compared to the standard ConvNets,
    \item address {\em a spectrum of different tasks} without ad-hoc manual tuning of receptive fields, 
    \item {\em eliminate the need for dilated convolutions} with  hand-crafted dilation factors, and
    \item {\em enable a novel analysis} of parameter allocation and spatial coverage in ConvNets.
\end{itemize}

This paper extends our (preliminary) work published in two conference papers~\citep{Tabernik2016a,Tabernik2018}, which considered only shallow architectures on small datasets. The DAU formulation is extended with details to cover the general form as well as its efficient formulation. Additional computer vision tasks are considered (i.e., blind image de-blur\-ring), significantly deeper architectures are used (Res\-Net50, Res\-Net101 and Deep\-Lab, SRN-Deblur\-Net) and several data\-sets (Citycape and GOPRO) are used to support the empirical findings. Implementations of DAU convolution filters in standard deep learning toolboxes\footnote{A low-level CUDA implementation of the DAU convolution filters are available in Caffe as well as Tensorflow at: \url{https://github.com/skokec/DAU-ConvNet-caffe} and \url{https://github.com/skokec/DAU-ConvNet}} are  available to the research community along with DAU modifications of popular deep architectures used in this paper. 

The remainder of the paper is structured as follows. Section~\ref{sec:related-work} provides a review of most closely related works. The DAU convolution filter is introduced in Section~\ref{sec:method}. A comprehensive empirical analysis of DAU convolution filter parameters and displacements is given in Section~\ref{sec:dau-analysis}. We demonstrate DAUs on standard computer vision tasks: classification (Section~\ref{sec:class-perf}), semantic segmentation (Section~\ref{sec:sem-segment}) and blind image de-blurring (Section~\ref{sec:deblur}). We conclude with a discussion in Section~\ref{sec:conclusion}.

\section{Related Work\label{sec:related-work}}

The receptive field has been considered as an important factor for deep networks in several related works~\citep{Luo2016,Chen2017}. \cite{Luo2016} measured an effective receptive field in convolutional neural networks and observed that it increases as the network learns. They suggest an architectural change that foregos a rectangular window of weights for sparsely connected units. However, they do not show how this can be implemented. Our proposed approach is in direct alignment with their suggested changes as our displaced aggregation units are a direct realization of their suggested sparsely connected units.

The importance of deforming filter units has also been indicated by recent work of \cite{Dai2017}. They implemented spatial deformation of features with deformable convolutional networks, while a general non-euclidean based formulation for a non-grid based input data was later proposed by \cite{Chang2018}. Both explicitly learn feature displacements but learn them on a per-pixel location basis for input activation maps and share them between all channels. \cite{Dai2017} also applies deformations only to the last few convolution layers on regions generated by Mask R-CNN \citep{He2017} with the effective outcome of normalizing scale-changes relative to the pixel position in the region of interest. Our model instead learns different displacements for different channels and shares them over all pixel locations in the input activation map. DAUs are applied to all layers with the goal of decoupling the receptive field size from the kernel size thus reducing the number of parameters and simplifying the architecture. Formulation of both methods also makes them conceptually complementary to each other.

Deforming filter units has also been explored by \cite{Jeon2017}, which, as opposed to deformable convolutions, apply deformation on filter units similarly as to our approach. They use bilinear interpolation similar to ours to get displacements at a sub-pixel accuracy, however, their limitation is in relying on $3\times3$ filters. They can neither displace them to more than a single neighboring pixel nor adapt them during the learning stage to an arbitrary position as we do. They also increase their parameter count as they still use 9 units per filter. We show that significantly fewer units are needed.

Works by \cite{Luan2017} and \cite{Jacobsen2016} changed the filter definition using different parametrization techniques.  Both decompose filter units into a linear combination of edge filters. They show a reduction in parameters per filter but their models do not allow displacements of filter units to arbitrary values. Their models have fixed receptive fields defined as hyperparameters and cannot be learned as ours. This also prevents any further analysis on distribution of displacements and receptive field sizes which is possible with our model with the explicit modeling of the unit's displacements.

Several papers also studied the importance of number of parameters. \cite{Eigen} performed an extensive evaluation and studied where parameters should be allocated, either for additional layers or for additional features. They concluded that it is more useful to allocate them for additional layers, but their study was limited by convolutional filter design and did not study how many should be allocated in spatial dimensions. Others have also observed inefficient allocation of parameters. For instance, \cite{Jaderberg2014} performed a low-rank analysis of pre-trained model weights and showed significant compression rate, while \cite{Iandola2016} proposed a more efficient network with 50-times less parameters. Most of the approaches reduce number of parameters through compression or architectural design change. Such compression techniques are complementary to ours and can be applied to our model as well.

\section{Displaced Aggregation Units (DAU)~\label{sec:method}}

We start by defining displaced aggregation units (DAUs) in their most general form. The derivatives required for learning in standard deep learning frameworks are presented in Section~\ref{sec:learning-daus} and an efficient formulation for fast inference is derived in Section~\ref{sec:efficent-dau}.

The activation map of the $i$-th feature (input into the current layer of neurons) is defined in the standard ConvNet as
 \begin{equation}
 	Y_{i} = f(\sum\nolimits_s W_{s}\ast X_{s}+b_{s}),
 \end{equation}
where for each $s$-th input channel, $b_s$ is a bias, $\ast$ is a convolution operation between the input map $X_s$ and the filter $W_{s}$, and $f(\cdot)$ is a non-linear function, such as ReLU or sigmoid~\citep{LeCun1998}. In convolution networks applied to images, the $X_s$ and $Y_i$ are two-dimensional features, while $W_s$ is two-dimensional convolution filter. We refer to the individual weight value of a filter $W_{s}$ in the standard ConvNet as its unit. 

\begin{figure}
\centering
\includegraphics[width=\columnwidth]{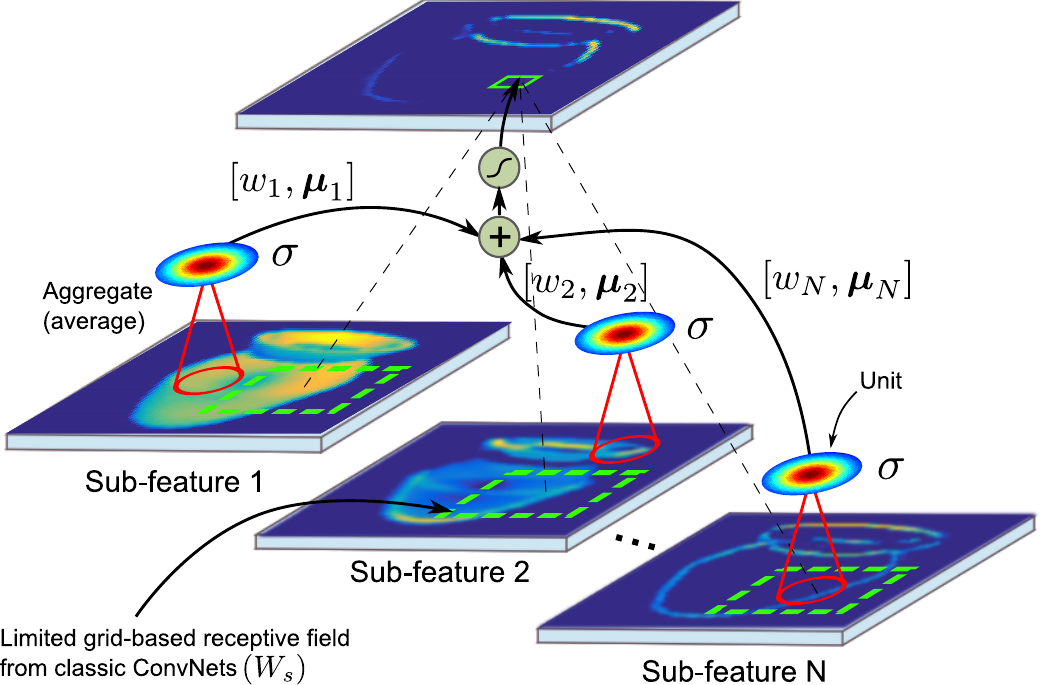}
\caption{Convolution filter with displaced aggregation units (DAUs) is composed of several displaced units that aggregate within a constrained area of the underlying sub-features from the lower layer.\label{fig:method}}
\end{figure}

We re-define the filters $W_{s}$ as a mixture of localized aggregated feature responses from the input feature map (see Figure~\ref{fig:method}). A Gaussian function is chosen for the analytic form of the aggregation units, although any other differentiable function that models the aggregation and displacement can be used. The resulting displaced aggregation unit (DAU) convolution filter is thus written as a mixture of $K$ units
 \begin{equation}
   W_{s}=\sum_{k=0}^K w_{k}G(\boldsymbol{\mu}_k, \sigma_k),\label{eq:weight-parametrization}
 \end{equation} 
where the unit's displacement and aggregation perimeter are specified by the mean $\boldsymbol{\mu}_k$ and standard deviation $\sigma_k$, respectively, and $w_{k}$ is the input amplification factor (i.e., the unit weight). Note that parameters $w_{k}$, $\boldsymbol{\mu}_k$ and $\sigma_k$ depend on the specific $s$-th input channel, as well as, on the specific $i$-th output channel, but in the interest of clarity we omit these in the notation. We refer to $\sigma_k$ as the aggregation perimeter since values of the Gaussian function at $3\sigma$ become small and its contribution will be negligible. Therefore, $3\sigma$ represents an approximate cutoff point of the unit's aggregation. 

The displaced aggregation unit, denoted by $G(\cdot)$, is implemented with a normalized Gaussian. To avoid discretization errors in $G(\boldsymbol{\mu}_k, \sigma_k)$ when implementing continuous function in a discrete convolution filter kernel, we replace the normalization factor computed in the continuous space with one computed in the discretized space, leading to our final aggregation unit $G(\vec{x}; \boldsymbol{\mu}_k, \sigma_k)$
 \begin{equation}
G(\vec{x};\boldsymbol{\mu_k}, \sigma_k)=\frac{1}{N(\boldsymbol{\mu}_k, \sigma_k)}\cdot \exp(-\frac{\left\Vert \vec{x}-\vec{\mu_k}\right\Vert^{2}}{2\sigma_k^{2}}),
\label{eq:gaussian-model}
\end{equation}
where $N(\boldsymbol{\mu}_k, \sigma_k)$ is the normalization term, i.e., 
 \begin{equation}
\begin{array}{c}
N(\boldsymbol{\mu}_k, \sigma_k)=\sum\nolimits_{\vec{x}}\exp(-\frac{\left\Vert \vec{x}-\vec{\mu_k}\right\Vert^{2}}{2\sigma_k^{2}}).
\end{array}
\end{equation}

The proposed DAU-based convolution filter is similar to a standard Gaussian mixture model, but we do not enforce $\sum w_k=1$ since the DAU weight $w_k \in[-\infty,\infty]$ can take any value. 

\subsection{Learning DAU Convolution Filter\label{sec:learning-daus}}

Learning the parameters of individual DAU consists of learning the displacement $\vec{\mu}_{k}$, the aggregation perimeter $\sigma_{k}$ and the weight $w_{k}$. The number of DAUs, $K$, in the convolution filter is a hyper-parameter that has to be set prior to learning. Since DAUs are analytic functions, the filter parameters are fully differentiable and compatible with the standard ConvNet gradient-descent learning techniques based on backpropagation. 

Parameters are thus optimized by computing the gradients w.r.t. the cost function $l(y,\bar{y})$, which leads to three different types of gradients. By applying the chain rule, we define the gradient for the weight $\frac{\partial l}{\partial w_{k}}$ as a dot-product of the back-propagated error and the input feature $X_{s}$ convolved with the $k$-th DAU, i.e.,
\begin{equation}
\frac{\partial l}{\partial w_{k}}=\underset{n,m}{\sum}\frac{\partial l}{\partial z}\cdot\frac{\partial z}{\partial{w}_{k}}=\underset{n,m}{\sum}\frac{\partial l}{\partial z}\cdot\underset{\vec{x}}{\sum}X_{s}\ast G(\vec{x};\boldsymbol{\mu}, \sigma),\label{eq:gradient-c-wrt-weight}
\end{equation}
where $n,m$ run over width and height of the image, $x$ runs over discretized kernel positions, while $z=\sum_s W_{s}\ast X_{s}+b_{s}$ and $\nicefrac{\partial l}{\partial z}$ is the back-propagated error. Note that only the $s$-th channel of input features are used since the weight ${w}_{k}$ appears only in $W_{s}$. The back-propagated error for layer $n$ follows the standard approach:
\begin{equation}
\frac{\partial l}{\partial z_{s}^{n}}=\frac{\partial l}{\partial z_{s}^{n+1}}\ast rot(W_{s}),
\end{equation}
where the back-propagated error from the higher layer $n+1$ is convolved with the 180\degree rotated filter $rot(W_{s})$ which can be computed from Eq.~(\ref{eq:weight-parametrization}). We can similarly apply the chain rule to obtain the gradient for the mean and the standard deviation,

\begin{equation}
\frac{\partial l}{\partial\mu_{k}}=\underset{n,m}{\sum}\frac{\partial l}{\partial z}\cdot \underset{\vec{x}}{\sum} X_{s}\ast\frac{\partial G(\vec{x};\boldsymbol{\mu_k}, \sigma_k)}{\partial\mu_{k}},\label{eq:gradient-c-wrt-mean}
\end{equation}

\begin{equation}
\frac{\partial l}{\partial\sigma_{k}}=\underset{n,m}{\sum}\frac{\partial l}{\partial z}\cdot \underset{\vec{x}}{\sum} X_{s}\ast\frac{\partial G(\vec{x};\boldsymbol{\mu}_k, \sigma_k)}{\partial\sigma_{k}},\label{eq:gradient-c-wrt-variance}
\end{equation}
where the derivatives of the Gaussian are 

{\small 
 \begin{equation}
\frac{\partial G(\vec{x};\boldsymbol{\mu}, \sigma)}{\partial\mu}={w}\frac{N(\boldsymbol{\mu}, \sigma)\cdot\frac{g(\vec{x};\boldsymbol{\mu}, \sigma)}{\partial\mu} - g(\vec{x},\theta)\cdot\frac{\partial N(\boldsymbol{\mu}, \sigma)}{\partial\mu}}{\left[N(\boldsymbol{\mu}, \sigma)\right]^{2}},
\end{equation}
}{\small \par}

{\small 
 \begin{equation}
\frac{\partial G(\vec{x},\boldsymbol{\mu}, \sigma)}{\partial\sigma}=\tilde{w}\frac{N(\boldsymbol{\mu}, \sigma)\cdot\frac{g(\vec{x})}{\partial\sigma}-g(\vec{x};\boldsymbol{\mu}, \sigma)\cdot\frac{\partial N(\boldsymbol{\mu}, \sigma)}{\partial\sigma}}{\left[N(\boldsymbol{\mu}, \sigma)\right]^{2}}.\label{eq:eq:gauss-wrt-variance}
\end{equation}
}
 
\subsection{Efficient Inference and Learning of DAUs\label{sec:efficent-dau}}

DAUs can be efficiently implemented in ConvNets by exploiting the translational invariance property of the convolution. The displacement of a Gaussian relative to the filter manifests in a shifted convolution result, i.e.,
\begin{align}	
    f \ast G(\boldsymbol{\mu}_{k},\sigma) &= f \ast \mathcal{T}_{\boldsymbol{\mu}_{k}}[G(\sigma)]\\
    & = \mathcal{T}_{\boldsymbol{\mu}_{k}}[f \ast G(\sigma)],
    \label{equ:gauss-trans-invariance}
\end{align}
where $\mathcal{T}_x(g,y) = g(y-x)$ is translation of function $g(\cdot)$ and $G(\sigma)$ is a zero-mean Gaussian. Thus, the activation map computation can be written as: 
\begin{align}
Y_{i}&=f\left(\underset{s}{\sum}\underset{k}{\sum}w_{k}\mathcal{T}_{\boldsymbol{\mu}_{k}}(G(\sigma) \ast X_{s})+b_{s}\right). \label{equ:fast-gauss-cnn}
\end{align} 
This formulation affords an efficient implementation by pre-computing convolutions of all inputs by a single Gaussian kernel, i.e., $\tilde{X}_{s}=G(\sigma) \ast X_{s}$, and applying displacements by $\boldsymbol{\mu}_{k}$ to compute the aggregated responses of each output neuron. The size of the blurring kernel is determined by the standard deviation ($2\cdot \left\lceil 3\sigma \right\rceil + 1$), however, large kernels do not add much computational cost since blurring represents only $1\% - 3\%$ of the whole computational cost. The efficient implementation requires sharing of the same aggregation perimeter $\sigma$ value among all units of the same layer. In a preliminary study, we have determined that this constraint is compensated for by the other free parameters in DAUs and performance is not affected. In fact, further constraints can be applied to the aggregation perimeters, which are empirically analyzed in Section~\ref{sec:variance-exp}. 

Due to discretization, the Eq.~(\ref{equ:fast-gauss-cnn}) is accurate only for discrete displacements $\boldsymbol{\mu}_{k}$. We address this by re-defining the translation function in Eq.~(\ref{equ:fast-gauss-cnn}) as a bilinear interpolation,
\begin{align}
\mathcal{\hat{T}}_x(g,y) &= \underset{i}{\sum}\underset{j}{\sum} a_{i,j} \cdot g(y - \left \lfloor{x}\right \rfloor + [i,j]),
\end{align}
where $a_{i,j}$ are bilinear interpolation weights. This allows computing sub-pixel displacements and can be efficiently implemented in CUDA kernels.

The aggregation perimeter constraints and the displacement re-formulation also make the learning more efficient. Only two parameters have to be trained per DAU, i.e., the weight $w_{k}$ and the spatial displacement ${\mu}_{k}$, while 
the aggregation perimeter and the number of DAUs per convolution filter are hyperparameters\footnote{Note that reasonable aggregation perimeter value $\sigma$ can in fact be estimated for a given problem by pre-training using the derivatives in Eq.~(\ref{eq:gradient-c-wrt-variance}), but using fixed value has proven sufficient. See Section~\ref{sec:variance-exp} for the analysis of different choices of this parameter.}.

Applying the efficient formulation to the learning of DAU convolution filter results in the following partial derivatives: 
\begin{align}
\frac{\partial l}{\partial w_{k}}&=\underset{n,m}{\sum}\frac{\partial l}{\partial z}\cdot\underset{\boldsymbol{x}}{\sum}\mathcal{\hat{T}}_{\boldsymbol{\mu}_{k}} (X_{s} \ast G(\sigma)),\\
\frac{\partial l}{\partial\mu_{k}}&=\underset{n,m}{\sum}\frac{\partial l}{\partial z}\cdot \underset{\boldsymbol{x}}{\sum} w_{k} \cdot \mathcal{\hat{T}}_{\boldsymbol{\mu}_{k}} (X_{s}\ast\frac{\partial G(\sigma)}{\partial\mu}),
\end{align}
where $\frac{\partial l}{\partial z}$ is back-propagated error. 

Similarly to the inference, the gradient can be efficiently computed using convolution with zero-mean Gaussian (or its derivatives) and sampling the response at displacement specified by the mean values in the DAUs.
 
The backpropagated error for the lower layer is computed similarly to the classic ConvNets, which convolve the backpropagated error on the layer output with rotated filters. Since the DAUs are rotation symmetric themselves, only the displacements have to be rotated about the origin and Eq.~(\ref{equ:fast-gauss-cnn}) can be applied for computing the back-propagated error as well, yielding efficient and fast computation.

\paragraph{Computational cost\label{sec:computational-cost}}

Compared to the implementation of DAUs based on standard convolution~\citep{Tabernik2016a} that discretize DAUs to large kernels, the efficient DAU implementation results in several times faster inference and an order of magnitude faster learning. However, the speed-up factor is dependent on the number of DAUs per channel $K$, and on the maximum displacement value. Considering the following input sizes:
\begin{align*}
X_s &= \left[W \times H \times S\right],~~ 
Y_i = \left[W \times H \times F\right],\\
G_k &= \left[\hat{\mathcal{K}}_w \times \hat{\mathcal{K}}_h \right],
\end{align*} where $S$ is the number of input channels, $F$ is the number of output channels and  $K$ is the number of DAUs per input channel, then the computational cost for the efficient DAU implementation is $\mathcal{O}(4 \cdot S \cdot F \cdot K \cdot W \cdot H + S \cdot W \cdot H \cdot \hat{\mathcal{K}}_w \cdot \hat{\mathcal{K}}_h)$, where $4$ relates to the bi-linear interpolation. Blurring with the Gaussian kernel $\hat{\mathcal{K}}_w \times \hat{\mathcal{K}}_h$ in the second term is performed $F$-times fewer than the first term, thus making the blurring part negligible for large number of input channels $F$. With the standard convolutional network using the $\hat{\mathcal{K}'}_w \cdot \hat{\mathcal{K}'}_h$ convolution kernel, the computational complexity is $\mathcal{O}(F \cdot S \cdot W \cdot H \cdot \hat{\mathcal{K}'}_w \cdot \hat{\mathcal{K}'}_h)$, and the speed-up factor $\gamma$ becomes:
\begin{align}
\gamma=\frac{\hat{\mathcal{K}'}_w \cdot \hat{\mathcal{K}'}_h}{4 \cdot K}.\label{equ:speedup}
\end{align}
With large DAU displacements resulting in bigger kernels for the standard convolution implementation, the speed-up of the efficient DAU implementation becomes more significant since kernel size $\hat{\mathcal{K}'}_w \cdot \hat{\mathcal{K}'}_h$ for the standard ConvNet must increase quadratically for larger displacements, while efficient DAU retains the same kernel size $\hat{\mathcal{K}}_w \cdot \hat{\mathcal{K}}_h$ and the same number of DAUs regardless of the displacements values.

\section{Analysis of the Displaced Aggregation Units}
\label{sec:dau-analysis}

An extensive empirical analysis was performed to gain insights into the properties of the displaced aggregation units CNN formulation (DAU-ConvNet). We have focused on two main parameters: (i) the DAU aggregation perimeter encoded by the variance of a Gaussian (Section~\ref{sec:variance-exp}) and (ii) the number of units per convolution filter (Section~\ref{sec:param-analysis}). We have then analyzed the learned convolution filters in terms of displacement distributions of the DAUs and thier receptive fields (Section~\ref{sec:adaptation-analysis}), and analyzed the practical computational savings (Section~\ref{sec:computational-cost-practice}).
 
\subsection{Influence of the DAU Aggregation Perimeter\label{sec:variance-exp}} 
\begin{table}
\centering
\small
\caption{Standard deviation $\sigma$ hyperparameter evaluation on CIFAR10 classification task using a shallow DAU-ConvNet. Standard deviation has minor effect on classification performance.}
\label{tab:variance-cifar}
\begin{adjustbox}{width=\columnwidth}

\begin{tabular}{lp{12pt}p{12pt}p{12pt}p{12pt}p{12pt}p{12pt}c}
\toprule
\textit{Std. deviation $\sigma=$} & $0.3$ & $0.4$ & $0.5$ & $0.6$ & $0.7$ & $0.8$ & Learned \\
\midrule
DAU-ConvNet  & \multirow{2}{*}{82.9} & \multirow{2}{*}{83.4} & \multirow{2}{*}{\textbf{83.8}} & \multirow{2}{*}{83.6} & \multirow{2}{*}{82.9} & \multirow{2}{*}{82.8} & \multirow{2}{*}{\textbf{84.25}} \\
 CIFAR10 & & & & & & & \\
\bottomrule
\end{tabular}
\end{adjustbox}
\end{table}
The aggregation perimeter of a single DAU is determined by the standard deviation, $\sigma$, of the corresponding Gaussian (Eq.~\ref{eq:gaussian-model}). In our most general formulation, the standard deviation can be learned for each unit by backprop (Eq.~\ref{eq:gradient-c-wrt-variance}), which in practice increases the computational complexity of the learning. The standard deviation plays several roles. On the one hand, it defines the region within which the DAU neuron aggregates features from a previous layer and on the other hand, it enables computation of smooth derivatives for DAU displacement optimization (Eq.~\ref{eq:eq:gauss-wrt-variance}) when $\sigma$ is large enough to encompass neigbooring pixels. We explore the trade-off between learning all the standard deviations and fixing them to a reasonable value that affords sufficient aggregation as well as displacement optimization.
\begin{figure*}
\centering
\includegraphics[width=1\textwidth]{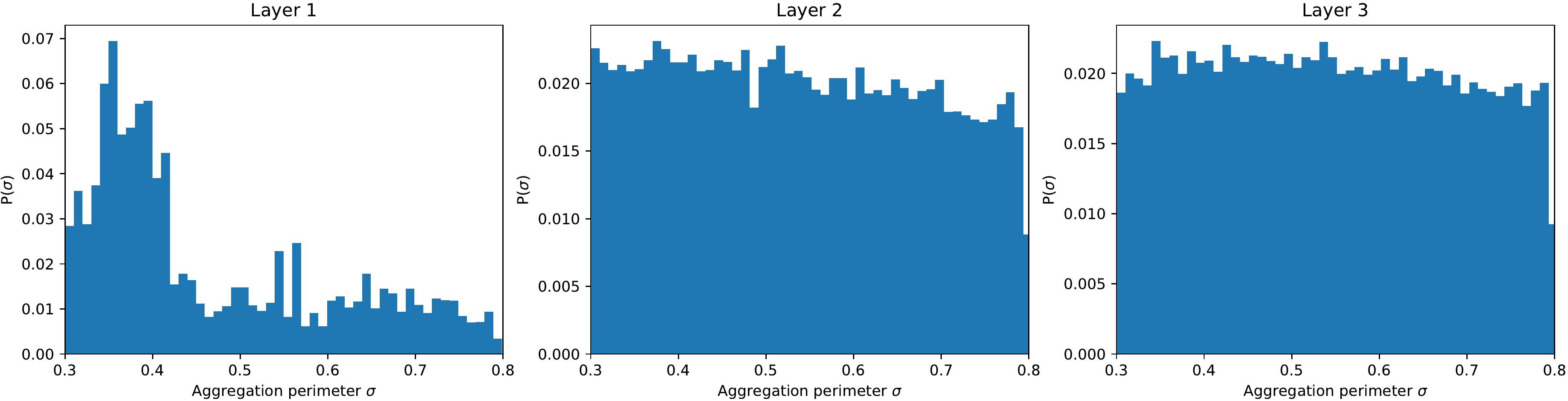} 
\includegraphics[width=1\textwidth]{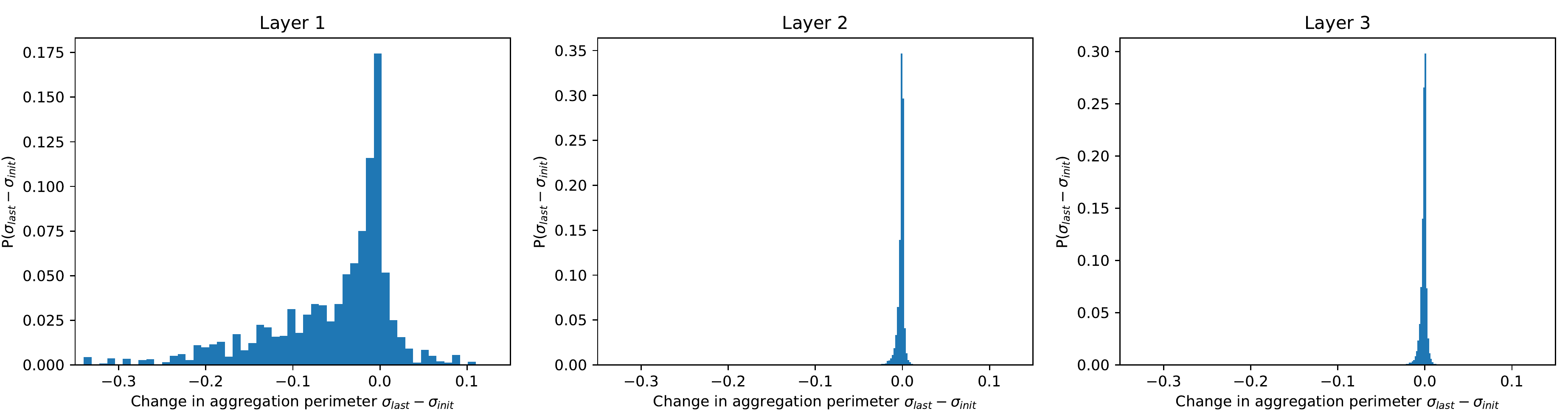} 
\caption{Distribution of learned DAU perimeters ($\sigma$) in the top row, and changes in the learned DAU perimeters from the initialization in the bottom row, both for model trained on the CIFAR10 dataset for each three layers in the network. 
\label{fig:learnable-variance-dist}}
\end{figure*}

The experiments were carried out on CIFAR10~ \citep{Krizhevsky2009} classification problem using a network with three convolutional layers with DAU filters and three max-pooling layers followed by a fully-connected layer for predicting the image class. Batch normalization~\citep{He2014} was applied and weights were initialized using~\citep{Glorot2010}. We trained the network with a stochastic gradient descent and  a softmax loss function for 100 epochs using a batch size of 256 images. Learning rate was set to 0.01 for the first 75 epochs and reduced to 0.001 for remaining epochs. A momentum of 0.9 was also used.

In the first experiment, the number of DAUs per convolution filter was fixed to four, while the standard deviations were learned. The standard deviations were initialized randomly with a uniform distribution over $[0.3,0.8]$. As reported in Table~\ref{tab:variance-cifar}, DAU-ConvNet with learned standard deviation achieved a 84.25\% accuracy. Top row in Figure~\ref{fig:learnable-variance-dist} shows the distribution of standard deviations for each layer after learning. Distributions for the second and the third layer remained uniformly distributed even after the training, while in the first layer, the standard deviations converged towards smaller values. The distributions in Figure~\ref{fig:learnable-variance-dist} were weighted by the learned unit's weight to reflect the changes of standard deviations only in the neurons that significantly contribute to the output. Comparison of the initial and final standard deviations in the bottom row of Figure~\ref{fig:learnable-variance-dist} confirms that learning affected only the first layer, while the following two layers were negligibly affected. This indicates that DAU structure does not benefit considerably from learning the standard deviations and supports the use of simplification that leads to efficient inference and learning.

As noted in Section~\ref{sec:efficent-dau}, the learning and inference of DAU-ConvNets can be made efficient by fixing the standard deviations in DAUs in the same layer. Table~\ref{tab:variance-cifar} also reports the results obtained by fixing the standard deviations to $\{0.3, 0.4,$ $0.5, 0.6, 0.7, 0.8\}$ in all layers. The results show that the classification rates vary by approximately 1\%, which means that the specific value of the perimeter negligibly affects the classification performance as long as it is set to a reasonable value. This means that unit's displacements and weights compensate for the chosen standard deviations. Comparing the performance of the DAU-ConvNet with the standard deviations fixed to $0.5$ and the DAU-ConvNet with the individually learned standard deviations (\textit{``Learned''} in Table~\ref{tab:variance-cifar}), we also observe a negligible difference of 0.5\%. Note that individually learned standard deviations prevent the use of efficient DAU implementation and require the implementation based on standard convolutions~\citep{Tabernik2016a}. Such implementation is significantly slower (cf. Section~\ref{sec:computational-cost}) and prevents the use of DAUs in very deep modern architectures.

Fixing the perimeters reduces the computational complexity of learning and very deep DAU-ConvNets can be trained. Thus in the remaining part of the experiments, we have fixed the DAU standard deviations to 0.5.

\subsection{Influence of the Number of DAUs \label{sec:param-analysis}}

With the standard deviations fixed, each DAU contains three parameters: two parameters for 2D displacement and a weight, which are learned from the data. A discrete parameter that has to be manually set, however, is the number of DAUs in the DAU convolution filter. This parameter thus determines the total number of parameters to be learned in the DAU-ConvNet and we analyze its impact on performance here using the ILSVRC 2012~\citep{Russakovsky2015} image classification task with a moderately deep standard ConvNet architecture.

\begin{table*}
\centering
\small{
\caption{Analysis of the number of parameters and units per filter with three variants of DAU-AlexNet: Large, Medium and Small. Rows also show the elimination of units based on their amplification value. In columns we report classification top-1 accuracy on ILSVRC2012 validation set, the number of DAU on all filters and the percentage of removed units. \label{tab:paramater-study}}

\begin{tabularx}{\textwidth}{Xccccccccc}
\toprule
\multirow{2}{*}{Relative threshold} & \multicolumn{3}{c}{\textit{Large DAU-AlexNet}} & \multicolumn{3}{c}{\textit{Medium DAU-AlexNet}} & \multicolumn{3}{c}{\textit{Small DAU-AlexNet}}\\
\cmidrule(l){2-4}\cmidrule(l){5-7}\cmidrule(l){8-10}
& Acc. (\%) & \# units & \% removed  & Acc. (\%) & \# units & \% removed & Acc. (\%) & \# units & \% removed \\ 
\midrule
0 & \textbf{57.3} & 1,523,712 & 0 & 56.9 &  786,432 & 0 & 56.4 & 393,216 & 0  \\
0.01 & \textbf{57.3} & 1,389,131 & 8  & 56.8 & 739,884 & 6 & 56.4 & 378,692 & 4 \\
0.02 & 57.1 & 1,325,057 & 13 & 56.7 & 707,745 & 10 & 56.4 & 366,144 & 7 \\
0.05 & 40.1 & 1,157,129 & 24 & 54.8 & 623,923 & 20 & 55.4 & 331,137 & 16 \\
0.10 & 28.3 & 925,509 &  39 & 47.4 & 507,651 & 35 & 49.6 & 279,162 & 29\\
0.25  & 0.2 & 453,987 & 70 & 1.9 & 261,093 & 66 & 0.9 & 154,624 & 61 \\
\bottomrule
\end{tabularx}
}
\end{table*}
In a classical ConvNet, the units are equivalent to pixels in the convolution filters. Several research papers investigated the influence of the number of parameters in classic ConvNets with respect to the number of layers, number of features and filter sizes~\citep{Eigen}, but could not analyze the impact of the number of units independently from the convolution filter size. The classic ConvNets are limited to a minimum of 9 parameters per filter, which corresponds to a $3\times3$ filter. Receptive fields may be increased with the dilated convolution~\citep{Holschneider1990} without increasing the number of parameters, but any such change requires hard-coding the size and the pattern, which leads to a combinatorial explosion of possible convolution filters.

The DAU formulation of convolution filter, on the other hand, allows us to investigate filters with even smaller number of parameters without affecting the spatial coverage and the receptive field sizes, since these are learned from the data. In addition, the DAU convolution filter definition (Eq.~\ref{eq:weight-parametrization}) provides a straight-forward way to prune the units. During the training, the number of units is kept the same in all filters. After the training is finished, the units with very small weights are removed, which further reduces the number of parameters.

\begin{table}
\centering
\small{
\caption{Per-filter unit and parameter count with three variants of DAU-ConvNet: Large, Medium and Small. Note, a unit in DAU has three parameters while a unit in a classic ConvNet has a single parameter.\label{tab:paramter-count}}

\begin{tabularx}{\columnwidth}{Xccccc}
\toprule
	& \multicolumn{5}{c}{\textit{Per-filter unit count}} \\
\cmidrule(l){2-6}
	& Large & Medium & Small & & AlexNet\\
\midrule
Layer 2 & 6 & 4 & 2 & &$5\times5$\\
Layers 3-5 & 4 & 2 & 1 & & $3\times3$\\
\midrule
	& \multicolumn{5}{c}{\textit{Per-filter parameter count}}  \\
\midrule
Layer 2 & 18 & 12 & 6 & &25 \\
Layers 3-5 & 12 & 6 & 3 & & 9 \\
\bottomrule
\end{tabularx}
}
\end{table}

\paragraph{Architecture.}
AlexNet~\citep{Krizhevsky2012}, composed of 7 layers (5 convolutional and 2 fully connected) was chosen for the baseline architecture for this experiment. We used a single-pipeline AlexNet, that does not require splitting into two streams as originally proposed~\citep{Krizhevsky2012}. For simplicity, we refer to a single-pipeline AlexNet as a standard AlexNet. The local normalization layers, max-pooling and dropout on fully-connected layers were kept. The initialization was changed to a technique by Glorot and Bengio~\citep{Glorot2010}. The baseline AlexNet was modified into a DAU-AlexNet by replacing discrete convolution filters in layers 2 to 5 with DAU convolution filters presented in Section~\ref{sec:method}. 

Three variations of DAU-AlexNet are constructed: Large, Medium and Small. Different number of units and parameters per kernel for each variant are shown in Table~\ref{tab:paramter-count} and follow approximate coverage of filter sizes from standard AlexNet with $5\times5$ filter sizes for the second layer and $3\times3$ filter sizes for the remaining layers. The Small DAU-AlexNet contains 400,000 DAUs, the Medium DAU-AlexNet contains 800,000 DAUs, and the Large DAU-AlexNet contains 1.5 million DAUs, which translates to 1.2 million, 2.3 million, and 4.5 million parameters, respectively. For reference, the baseline AlexNet contains 3.7 million parameters.

\paragraph{Dataset.}

The networks were trained on 1.2 million training images from ILSVRC 2012~\citep{Russakovsky2015} and tested on 50,000 validation images. All images were cropped and resized to 227 pixels as in the reference AlexNet architecture~\citep{Krizhevsky2012}. To keep the experiments as clean as possible, we did not apply any advanced augmentation techniques apart from mirroring during the training with a probability of $0.5$.

\paragraph{Optimization.}

The networks were trained by stochastic gradient descent with the batch size of 128 for 800,000 iterations, or 80 epochs. The initial learning rate was set to 0.01 and was reduced by a factor of 10 every 200,000th iteration. A momentum with a factor of 0.9 was applied together with a weight decay factor of 0.0005. The weight decay was applied only on the weights $w_k$ in DAUs but not to the displacement values $\boldsymbol{\mu_k}$.

\paragraph{Results and discussion.}

The results are reported in Table~\ref{tab:paramater-study}. We observe that all three DAU-AlexNets achieve classification accuracy of approximately 56-57\%, which is comparable to the standard AlexNet (cf. Section~\ref{sec:class-perf}). This performance is already achieved by a DAU-AlexNet with two or less units per convolution filter, resulting in 3 to 6 parameters per filter, which is significantly lower than in a standard AlexNet with  9 parameters for the smallest filter (i.e., $3 \times 3$) and 25 for a moderately large filter (i.e., $5 \times 5$). Note that the efficient use of parameters is possible since DAU-AlexNet learns the required convolution filter receptive field size through its adjustable units without increasing its parameters.

\begin{figure}
\centering
\includegraphics[width=0.9\columnwidth]{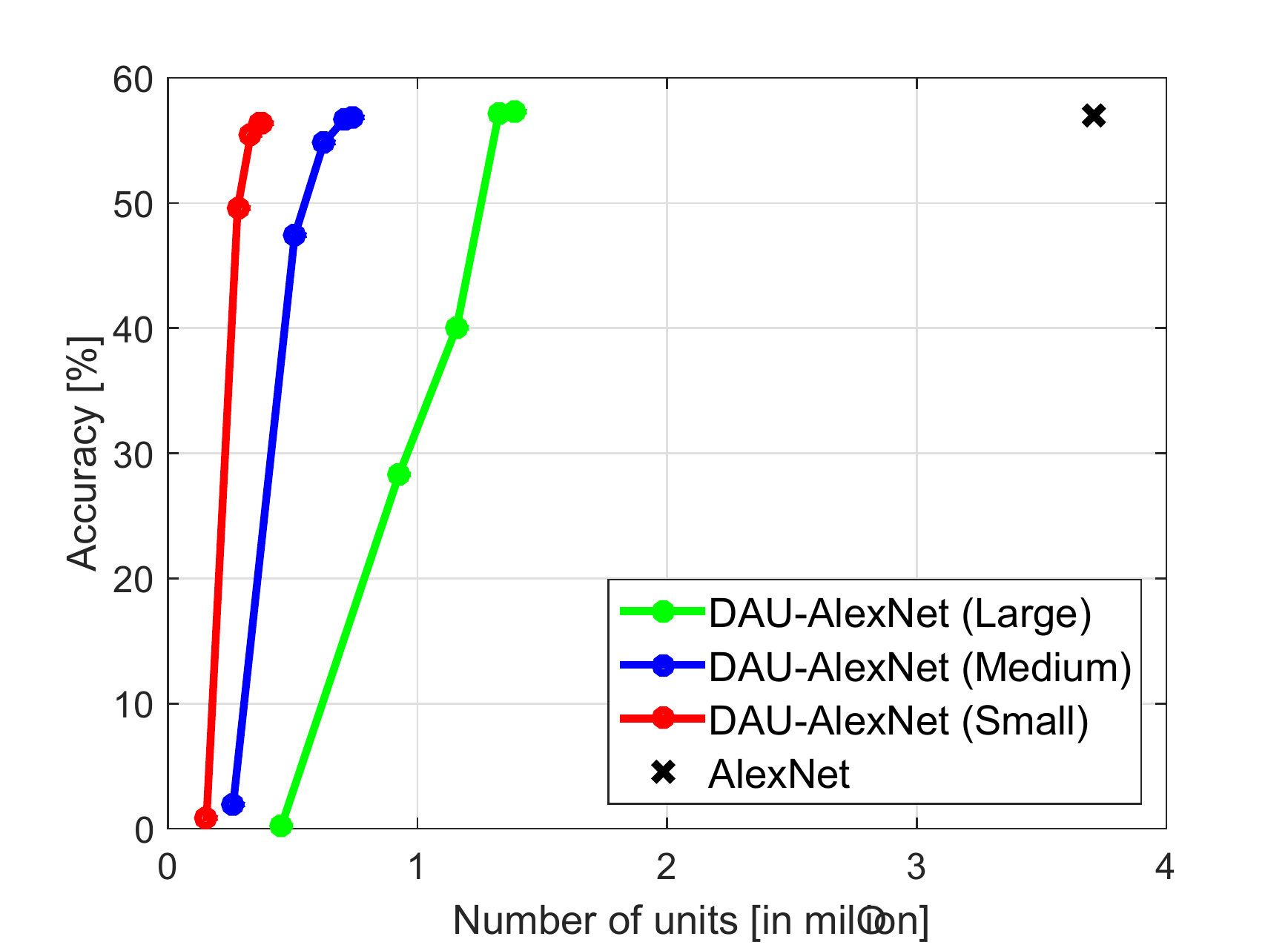}
\caption{Classification accuracy on ILSVRC 2012 dataset with respect to the number of DAUs in the network (in million). DAU-based AlexNet requires an order of magnitude fewer units than standard AlexNet at the same classification accuracy. Note that in DAU-AlexNet, the unit refers to one Gaussian component with a learnable weight and displacement, while in AlexNet, the unit refers to one weight in a kernel. \label{fig:acc-vs-units-alexnet}}
\end{figure}
Further improvements are observed after the pruning -- eliminating units with low weights. Table~\ref{tab:paramater-study} shows that in all DAU-AlexNets, 7-13\% of units can be removed without reducing the classification accuracy. The relation between the number of parameters and the classification performance is visualized in Figure~\ref{fig:acc-vs-units-alexnet}. A steep increase in performance by the Small DAU-AlexNet shows that even pruning the smaller network can reduce some parameters while retaining fairly good results. Almost 50\% accuracy can be maintained with less than 300,000 units, which is 10-fold less than in standard AlexNet. Furthermore, comparing to the less steep decline of the Large DAU-AlexNet shows that pruning is less effective than just learning with a limited set of units. This indicates that larger network has encoded its information over many more units than are necessary.

\subsection{Spatial Adaptation of DAUs \label{sec:adaptation-analysis}}

Next, we have investigated spatial distribution of the learned DAUs displacements from the convolution filter centers. The aim of the experiment was to expose whether certain displacement sizes are favored for a given task, and what are the indicated receptive field sizes of the convolution filters.

\begin{figure*}
\centering
\includegraphics[width=\textwidth]{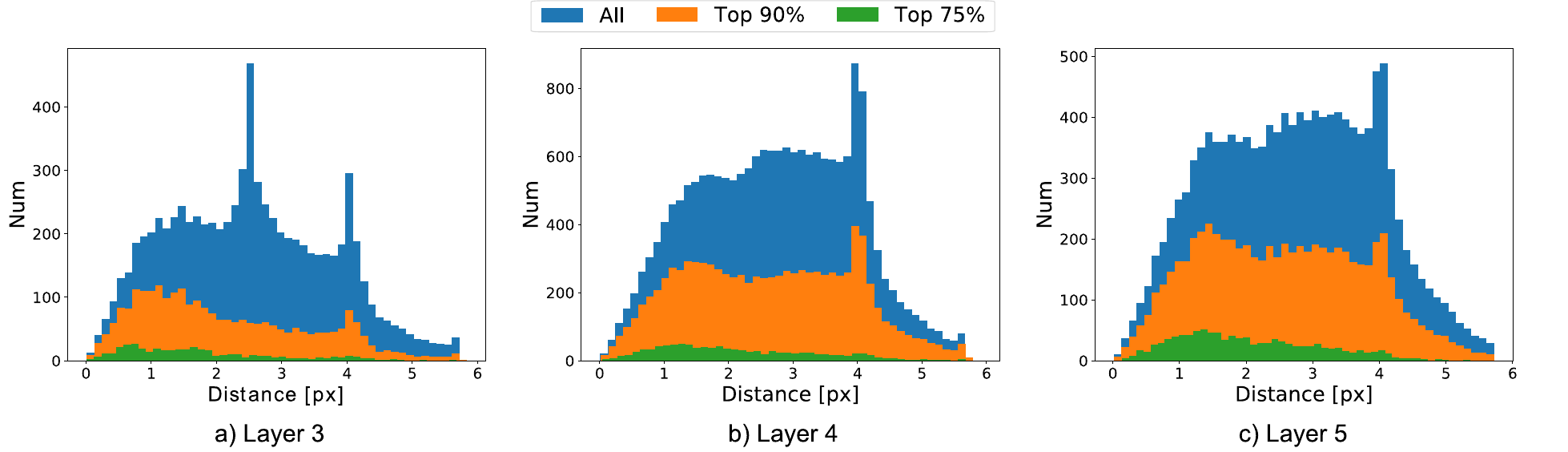}
\caption{Distance-to-center distributions collected from displacement of DAUs. Distributions are shown per-layer and after keeping only units corresponding to the top percentage of absolute weights (blue, orange, green).\label{fig:spatial-dist-1d}}
\end{figure*}

Such an experiment is practically unfeasible with classical ConvNets and requires a combinatorial sweep over alternative architectures with various hand-crafted filter designs. On the task of segmentation, for example, convolution filter receptive fields may be increased by dilated convolutions~\citep{Chen2016a}, but the dilation factor has to be manually set. In contrast, the DAU convolution filters optimize their units with sub-pixel accuracy and can vary across the filters, thus no hand-crafting is required.

We investigate a 1D distribution of distances to the convolution filter centers as well as 2D distributions aggregated over all convolution filters. In both distributions a specific DAU contributes to the overall distribution proportionally to the unit absolute weight.

\paragraph{Architecture.}
A pre-trained Medium DAU-Alexnet architecture from the previous section was adapted to the segmentation task to perform a fine pixel-level class prediction as follows. The last fully-connected classification layer was replaced by the expansion and classification layer from \cite{Long2015} that entails a $1\times1$ classification layer and bi-linear up-sampling with deconvolution layer to obtain pixel-wise mask. By removing the last two max-pooling layers we further increase the resolution which results in object boundaries maintained sharp. In this way the down-sampling factor is reduced from 32$\times$ to 8$\times$. Increasing the resolution of a pre-trained DAU-AlexNet model results in missaligned DAU positions, which were trained for a lower resolution. This is compensated for by proportionally increasing the displacements of DAUs in the layers with the increased resolution.

\paragraph{Dataset.}
The PASCAL VOC 2011~\citep{pascal-voc-2011} segmentation challenge was used. The training set was a combination of 1,112 training images from the PASCAL VOC 2011 segmentation challenge and 7,386 images collected by \cite{Hariharan2011}. Performance was evaluated on the PASCAL VOC 2011 validation set excluding the images from \cite{Hariharan2011}.

\paragraph{Optimization.}
The models were trained by mini-batch stochastic gradient descent for 65,000 iterations (150 epochs) with a batch size of 20 images. A fixed learning rate of 0.0002 was used, weight decay was set to 0.0005 and momentum to 0.9. Similar to~\cite{Long2015}, the added classification layer was initialized with zeros 
and a normalized per-pixel softmax loss was applied on pixels with valid labels.
 
\begin{figure}
\includegraphics[width=\columnwidth]{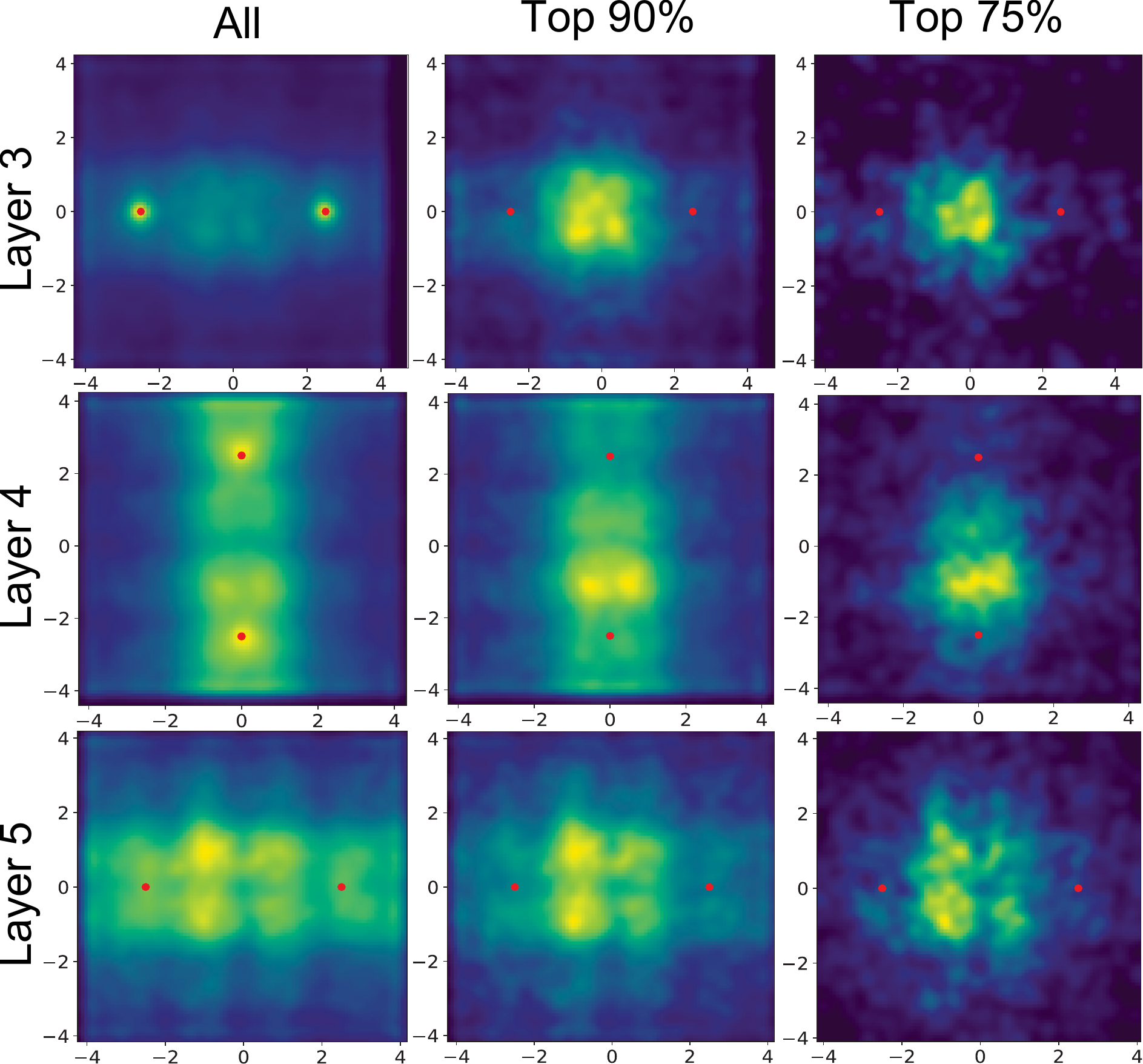}
\caption{2D distributions of displacements collected from DAUs. Red dots indicate initialization points. Distributions reported for layer 3, 4 and 5 in top, middle and bottom row, respectively. The three columns show the distributions after keeping only units corresponding to the top percentage of absolute weights.\label{fig:spatial-dist-2d}}
\end{figure}

\paragraph{Results and discussion.}
The DAU-AlexNet achieves seg\-men\-ta\-tion accuracy comparable to the standard AlexNet with dilated convolutions (c.f. Section~\ref{sec:sem-segment}), which verifies that the network has properly adapted to the task. DAU displacement distributions were computed separately for layers 3, 4 and 5. In particular, three different distributions were computed. The first distribution considered locations of all DAUs, the second considered locations of the DAUs with weight at least $90\%$ of the largest absolute weight and the third considered locations of the DAUs with the weight at least $75\%$ of the largest absolute weight. The resulting 1D and 2D distributions are visualized in Figure~\ref{fig:spatial-dist-1d} and Figure~\ref{fig:spatial-dist-2d}, respectively.

Two significant spikes are observed in the 1D distributions in Figure~\ref{fig:spatial-dist-1d}. One spike corresponds to 2.5 pixels displacement and the other to 4 pixels displacement. The spike at 2.5 pixels occurs only at the third layer and corresponds to DAU initialization points, which means that many units did not move significantly. This is confirmed by the high density regions in 2D distributions Figure~\ref{fig:spatial-dist-2d}, at initialization centers (red dots). However, a further inspection reveals that these units do not contribute in the inference, since their weights are very small. In fact, they disappear in the distributions of Figure~\ref{fig:spatial-dist-1d} and Figure~\ref{fig:spatial-dist-2d} with the DAUs corresponding to negligible weights removed. This is in effect the self-pruning property of DAU-ConvNets which was observed in Section~\ref{sec:param-analysis}. The spikes at initialization points are not apparent at the 4th and the 5th layers in the corresponding distributions (Figure~\ref{fig:spatial-dist-1d} and Figure~\ref{fig:spatial-dist-2d}). This means that in this particular problem the DAUs are redundant only at the 3rd layer.

The second spike at 4 pixels is significant and does not disappear when removing DAUs with small weights (Figure~\ref{fig:spatial-dist-1d}). The spike occurs due to a limitation of our implementation that constraints the receptive field size, which in our case is set at four pixels in both spatial dimensions\footnote{Our current implementation in CUDA allows only distances up to 4 or 8 pixels. This limitation can be overcome by modifying the implementation.}. Still, a significant number of those units have large weights, which suggests that even larger receptive fields would be observed if unconstrained by the implementation specifics of GPU processing.
 
The overall shape of the displacement distribution is consistent across all layers (Figure~\ref{fig:spatial-dist-1d}). This indicates a preference to densely cover locations 1-2 pixels away from the center for the segmentation task. Some units with large weights are located far away from the center, which indicates a need to cover large receptive fields albeit with a lower density. The same conclusion is drawn from 2D spatial distributions in Figure~\ref{fig:spatial-dist-2d}.

\begin{figure}
\centering
\includegraphics[width=\columnwidth]{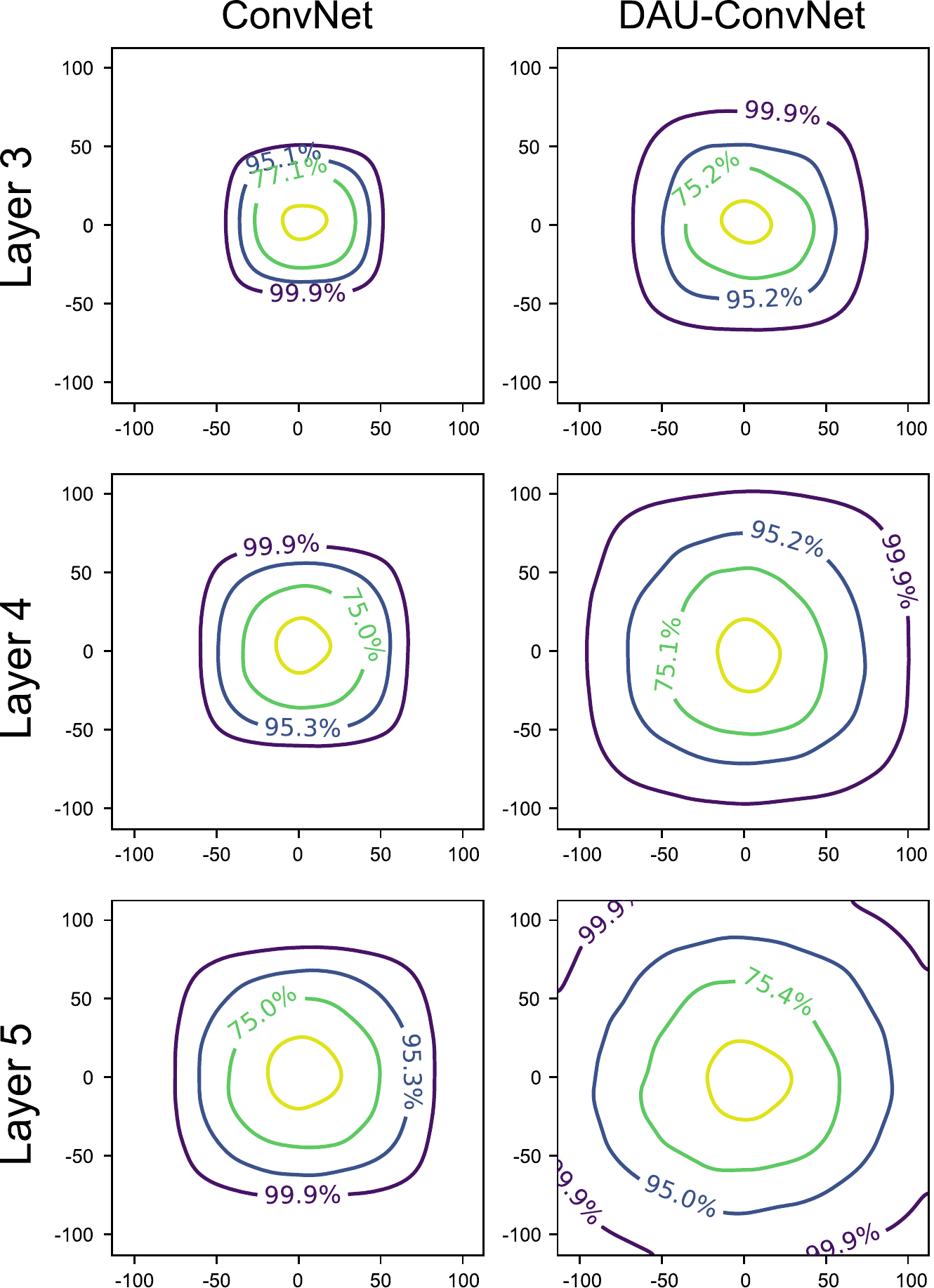}
\caption{The effective receptive fields (ERF) in a contour plot visualization for layers 3, 4 and 5 in standard AlexNet (left) and DAU-AlexNet (right) trained for the semantic segmentation. The size of the visualization patch is $227\times227$ pixels. Note the inner (yellow) contour presents a 25\% influence area. \label{fig:erf}}
\end{figure}
\paragraph{The Effective Receptive Field.}

We further visualize the receptive field of the network by calculating the Effective Receptive Field (ERF) as introduced by~\cite{Luo2017}. The ERF measure calculates the effective receptive field of a single output pixel in a specific channel by back-propagating the error with only one active pixel in the corresponding channel. We report ERF averaged over all the channels for each layer and depict ERF as contour plots that well capture the extent of the receptive field. A contour represents the area with a fixed percentage of the influence to the output neuron. For instance, all pixels within 75\% line represent 75\% of the whole influence on the output neuron. 

The effective receptive fields for DAU-AlexNet and standard AlexNet trained on the semantic segmentation are reported in Fig.~\ref{fig:erf}. In all three layers, the DAU-ConvNet consistently demonstrates larger receptive field sizes than the ConvNet. The most noticeable difference is demonstrated in the 5th layer in the bottom row, where in the standard ConvNet the 99.9\% of the influence to the output neuron is concentrated at only approximately 60 pixels from the center, while in DAU-CovNet, this influence is extended by nearly a factor of two.

\subsection{Computational cost} \label{sec:computational-cost-practice}

As show in Section~\ref{sec:computational-cost}, the computational cost of the DAU model is dependent only on the number of DAUs per channel, and not on the size of the convolution kernel, as in the standard convolution. In practice, we found $K=2$ and 4 pixel displacements (corresponding to $\hat{\mathcal{K}'}_w \cdot \hat{\mathcal{K}'}_h=9\times9$) are sufficient for large networks such as AlexNet or ResNet. This results in a theoretical speed-up of $\gamma=10.125$. Profiling an efficient DAU implementation on a shallow 3-layer architecture with $K=2$, $\hat{\mathcal{K}'}_w \cdot \hat{\mathcal{K}'}_h=9\times9$, $192$ output features and $32\times32$ input resulted in $3.25$ times faster inference and $12.54$ times faster learning on NVIDIA RTX 2080 Ti compared to the implementation based on the standard convolution. Note that in the efficient DAU the sigma was not learned, therefore adding additional 1.33-times speed-up compared to the theoretical speed-up (Eq.~\ref{equ:speedup}), while the difference to the theoretical speed-up for the inference points to the overhead cost and inefficiencies in our CUDA implementation compared to the CuDNN implementation.

\begin{table}
\centering
\caption{Results on ILSVRC 2012 validation set using AlexNet architecture and corresponding number of parameters on convolutional layers. Top-1 accuracy reported. \label{tab:cls-results}}
\begin{tabularx}{\columnwidth}{Xccc}
\toprule
 \textit{Network architecture} &  \makecell[c]{Top-1\\accuracy (\%)} & \makecell[c]{Number of parameters\\on conv. layers}  \\ 
\midrule
DAU-AlexNet & 56.89  &  \textbf{2.3} million  \\
AlexNet & \textbf{56.99} & 3.7 million \\
\bottomrule
\end{tabularx}
\end{table}

\section{Application to Classification\label{sec:class-perf}}

The generality of DAU-ConvNets is demonstrated on several computer vision tasks with state-of-the-art CNN architectures. In this section, DAU-ConvNets are empirically analyzed on the ILSVRC 2012 classification task using the AlexNet and ResNet models, while the following sections (Section~\ref{sec:sem-segment} and  Section~\ref{sec:deblur}) demonstrate application to semantic segmentation and de-blurring.

\subsection{AlexNet with DAUs}

The first experiment involved evaluation on a classic architecture AlexNet~\citep{Krizhevsky2012} from Section~\ref{sec:param-analysis}. We compare the baseline AlexNet to Medium DAU-AlexNet (Table~\ref{tab:paramter-count}), which contains less than 70\% of parameters than the baseline.

Table~\ref{tab:cls-results} reports accuracy for both methods together with the number of free parameters in the convolution layers. The DAU-AlexNet and the baseline AlexNet converge to comparable performance, close to 57\%. The DAU-version of AlexNet achieved comparable performance to the classical  Alex\-Net with over 30\% fewer parameters and analysis in Section~\ref{sec:param-analysis} shows that further reduction is possible at negligible performance loss. The overall comparable performance supports the hypothesis that DAUs do not lose expressive power on the account of their simple functional form.

\begin{table*}
\small{
\caption{Results on segmentation task using a PASCAL VOC 2011 validation set. Per-class mean-IU and averaged mean-IU over all classes are reported.  \label{tab:voc2011-results}}
\begin{adjustbox}{width=1\textwidth}
\begin{tabular}{ 
lp{0.3cm}p{0.3cm}p{0.3cm}p{0.3cm}p{0.3cm}p{0.3cm}p{0.3cm}p{0.3cm}p{0.3cm}p{0.3cm}p{0.3cm}p{0.3cm}p{0.3cm}p{0.3cm}p{0.3cm}p{0.3cm}p{0.3cm}p{0.3cm}p{0.3cm}p{0.3cm}cc }
\toprule
\textit{Network arch.}& \rot{bg} & \rot{arpln.} & \rot{bcyle.} & \rot{bird} & \rot{boat} & \rot{bottle} & \rot{bus} & \rot{car} & \rot{cat} & \rot{chair} & \rot{cow} & \rot{d.tble} & \rot{dog} & \rot{horse} & \rot{m.bike} & \rot{person} & \rot{p.plant} & \rot{sheep} & \rot{sofa} & \rot{train}&\rot{tv.} & mIoU\\
\midrule
DAU-AlexNet & \textbf{86.1} & \textbf{58.5} & \textbf{29.7}& \textbf{55.0} & \textbf{41.7} & \textbf{47.2} & \textbf{61.3} & \textbf{56.3}& \textbf{57.9} & \textbf{14.1} & \textbf{47.1} & \textbf{27.3} & 47.8 & \textbf{36.7} & \textbf{54.7} & \textbf{63.9} & \textbf{28.9} & 53.0 & \textbf{19.3} & 59.8 & \textbf{45.3} & \textbf{47.22}\\
AlexNet-dilation & 85.8 & 54.6 & 27.2 & 51.8 & 39.0 & 45.2 & 56.3 & 54.2 & 57.4 & 12.4 & 43.8 & 26.1 & \textbf{50.6} & 35.6 & 54.1 & 61.1 & 26.9 & \textbf{53.6} & 18.9 & \textbf{60.2} & 42.5 & 45.57\\ 
\bottomrule
\end{tabular}

\end{adjustbox}
}
\end{table*}
\subsection{Residual Networks with Displaced Aggregation Units}

Next, we evaluated DAUs on ResNet50 and ResNet101~\citep{He2015a} classification architectures. In particular, we evaluated ResNet~v2~\citep{He2016}, which applies batch normalization and activation before convolution for learning stabilization.

ResNet was modified into a DAU-ResNet by replacing all $3\times3$ convolutions with DAU convolution filters containing only two units (see Figure~\ref{fig:intro-resnet}). This includes all layers except the first layer with $7\times7$ kernels and bottleneck layers with $1\times1$ kernels. We also implemented down-sampling with max-pooling instead of using convolutions with a stride\footnote{DAU layers with stride operation are not yet implemented.}. This was performed on all levels except on the first one, where standard convolution was retained. The same down-sampling with max-pooling was included in the standard ResNet for a fair comparison. In DAU-ResNet, the displaced aggregation units were initialized randomly with uniform distribution on interval $[-1.5,1.5]$, following the observation of displacement distribution in Section~\ref{sec:adaptation-analysis}. Units were restricted to move up to 4 pixels away from the center, resulting in receptive field size of up to $9\times9$ pixels relative to the previous layer. This restriction was enforced only due to technical limitations in current DAU implementation. 

\paragraph{Optimization.}
Both architectures were trained by stochastic gradient descent. The same optimization hyper-parameters were used in DAU and classic ResNet, i.e.,  learning rate of $0.1$, momentum of $0.9$, weight decay of $10^{-4}$ and a batch size of 256. Learning rate was reduced four times by a factor of 10 at 30\textsuperscript{th}, 60\textsuperscript{th}, 80\textsuperscript{th} and 90\textsuperscript{th} epoch. In DAUs, the weight decay was applied only to weights but not to offset; however, 500-times larger learning rate for offset was used during the training to compensate for orders of magnitude different values compared to the weights. 

\paragraph{Classification Results with ResNet.}
Results for networks with 50 and 101 layers are reported in Table~\ref{tab:cls-results-resnet}. The DAU version achieves the same performance as the classical ConvNet counterparts on ResNet50 as well as ResNet101. This result is achieved with a 30\% reduction of parameters allocated for convolutions in spatial coverage of DAU-ResNet. The reduction in the overall number of parameters is slightly lower since the residual network allocates half of the parameters for $1\times1$ bottleneck layers, which are not replaced with DAUs.

\section{Application to Semantic Segmentation\label{sec:sem-segment}}

\begin{table}
\centering
\caption{Results on ILSVRC 2012 validation set using deep residual network architecture and corresponding number of parameters on convolutional layers. Top-1 accuracy reported. \label{tab:cls-results-resnet}}
\begin{tabularx}{\columnwidth}{Xccccc}
\toprule
 \multirow{2}{*}{\textit{Network arch.}} &  \multirow{2}{*}{\makecell[c]{Top-1\\acc. (\%)}}& \multicolumn{3}{c}{Number of parameters (in million)}  \\ 
 \cmidrule{3-5}
 &   & Conv/DAU & Bottlenecks & Total \\
\midrule
ResNet50 & \textbf{74.08} & 11.3 M & 14.2 M & 25.5 M \\
DAU-ResNet50 & 74.06 & \textbf{7.5} M & 14.2 M & \textbf{21.7} M\\
\midrule
ResNet101 & \textbf{75.39} & 21.3 M & 23.1 M & 44.5 M\\
DAU-ResNet101 & 74.89 & \textbf{14.2} M & 23.1 M & \textbf{37.4} M \\
\bottomrule
\end{tabularx}
\end{table}
Classic ConvNet architectures designed for classification require hand-crafted structural modifications in the form of hand-tuned dilated convolutions to achieve high-quality results on other task like semantic segmentation. In particular, dilated convolution with several manually set dilation sizes are placed at certain layers in the ResNet when adapting for semantic segmentation~\citep{Chen2014}. Such changes are not required for DAU counterparts, since these simultaneously learn the filter receptive field sizes and content to the task at hand. A semantic segmentation task on PASCAL VOC 2011 and Cityscape datasets using three popular deep learning architectures, AlexNet, ResNet101 and DeepLab was chosen to demonstrate this.

\begin{figure*}
\includegraphics[width=\linewidth]{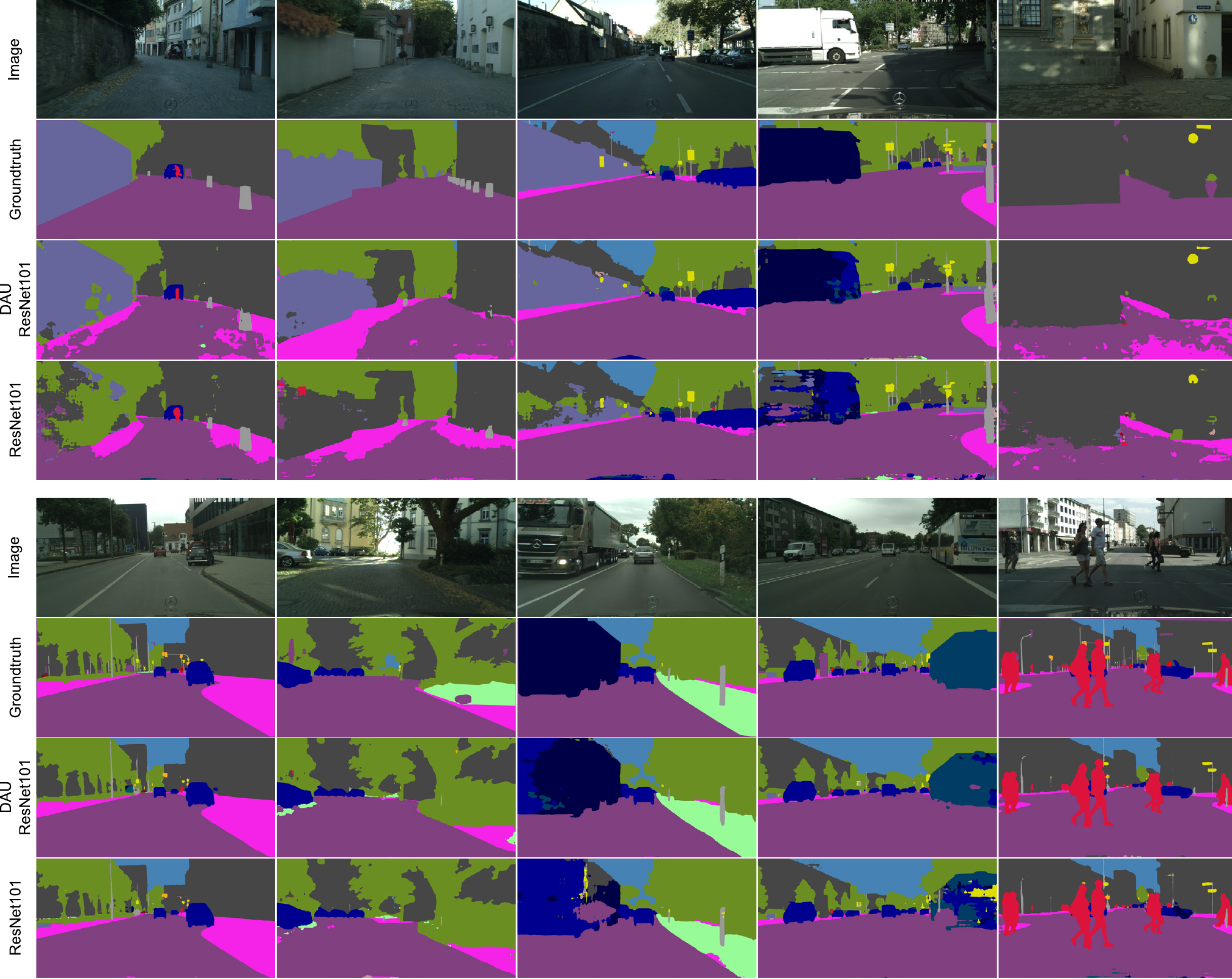}
\caption{Examples of semantic segmentation on Cityscape dataset with DAU-ResNet101 in the third and seventh row and standard ResNet101 fourth and eigth row.\label{fig:cityscape-examples}}
\end{figure*}

\subsection{Semantic Segmentation with AlexNet}

We first evaluated a classic architecture -- the AlexNet model. An AlexNet architecture modified for semantic segmentation from Section~\ref{sec:adaptation-analysis} was evaluated on PASCAL VOC 2011 segmentation dataset. Modification includes increased resolution at the last two layers and scaled displacements in the corresponding DAU convolution filters. The baseline AlexNet was similarly modified for the semantic segmentation task, but instead of scaling the displacements, we dilated convolution filters with the same factor. The layers after the first removed max-pooling use a dilation of two (layers 3, 4 and 5) and the layers after the second-removed-max-pooling use a dilation factor of four (layers 6 and 7). 

\paragraph{Segmentation results.}
The performance of DAU-AlexNet compared to the baseline AlexNet with dilation is shown in Table~\ref{tab:voc2011-results}. DAU-AlexNet consistently outperforms the baseline AlexNet with dilation across all measures. The mean IoU and per-pixel accuracy are improved by approximately 2\%. Looking at the per-class mean IU, we observe the improvement is consistent over all categories, with the exception of "dog", "sheep" and "train". 
 
\subsection{Semantic Segmentation with Residual Networks}

DAUs were further evaluated on a very deep residual network with 101 layers (ResNet101)~\citep{He2015a}. Res\-Net101 was modified in the same way as AlexNet in the previous section. This included removal of the last max-pooling layer, which resulted in a network with output layer resolution reduced by 16$\times$ (as opposed to 32$\times$ reduction in the original ResNet101). This matches to having the output stride of $16$ as in the DeepLab model~\citep{Chen2014}.

\begin{table*}
\centering
\caption{Results on Cityscape validation set using deep residual network architecture and DeepLabv3+ improvements. We report mean intersection-over-union (mIoU). DAU-6U is a single displaced aggregation layer with 6 units per channel which replaces ASPP. \label{tab:seg-results-resnet}}
\begin{tabularx}{\linewidth}{cccccXccccc}
\toprule
\multicolumn{5}{c}{\textit{Standard ResNet101 backbone}} & &
\multicolumn{5}{c}{\textit{ResNet101 with DAUs (our) backbone}} \\
\cmidrule{1-5} \cmidrule{7-11}
  Output Stride & ASPP & Image-pool & Decoder & mIoU & \textit{+/-} & mIoU & Output Stride & DAU-6U & Image-pool & DAU-Decoder\\
\midrule
16 & & & & 68.6 & \textit{+4.2} & \textbf{72.8} & 16 & & &  \\
\midrule
16 &  & \checkmark & & 72.7 & \textit{+0.1} & \textbf{72.8} & 16 & & &  \\
16 & \checkmark & \checkmark & & \textbf{75.6} & \textit{-0.1} & 75.5 & 16 & \checkmark & \checkmark & \\
16 & \checkmark & \checkmark & \checkmark & 75.8 & \textit{+0.3} & \textbf{76.1} & 16 & \checkmark &  \checkmark & \checkmark \\
\bottomrule
\end{tabularx}
\end{table*}

\paragraph{Cityscape dataset.}
The Cityscape dataset~\citep{Cordts2016} was used for evaluation. The dataset contains high-resolution images of city driving and the task requires pixel-wise segmentation of the image into 19 classes. Only fine-grained annotations were used (i.e., 2,975 training images) and the networks were evaluated on 500 test images from the validation set.
 
\paragraph{Optimization.}
The standard ResNet and DAU-ResNet were first pre-trained on ImageNet~\citep{Russakovsky2015}. Both models were then trained for segmentation using a mini-batch stochastic gradient descent with a batch size of 8 for 50,000 iterations (134 epochs). A learning rate of 0.01 was used, with momentum of 0.9 and a weight decay of $10^{-4}$. A polynomial decay of the learning rate with a factor of 0.9 was applied. Data augmentation was used with the following operations: images were resized by a factor randomly selected from a uniform distribution in a range of $[0.5 , 2.0]$ and high-resolution images were randomly cropped into $769\times769$ large patches, and left-to-right mirroring was applied with a probability of 0.5. Testing was performed on a single-scale without multi-scale testing.

\paragraph{Results.}
Results are reported in the first row in Table~\ref{tab:seg-results-resnet}. DAU-ResNet101 achieves 72.8\% mIoU and outperforms the standard ResNet101 (68.6\% mIoU). A similar difference is observed in the mean accuracy -- standard ResNet101 obtains 77.2\% mean accuracy, while the DAU version obtains a 82.2\% mean accuracy. Improvement of around 4\% clearly demonstrates the benefits of having learnable unit displacements, which allow the network to focus on spatial features required for segmentation without requiring manual specification. Note that a 4\% increase in the performance was achieved with 15\% less parameters in the network. Several examples of semantic segmentation on both networks are depicted in Figure~\ref{fig:cityscape-examples}. Several top examples  well demonstrate gridding artifacts in ResNet101 while DAUs avoid this issue.

\subsection{Improving DeepLab with DAUs}

Since DAUs inherently provide adjustable receptive field sizes, it becomes a natural fit for a popular semantic segmentation model, DeepLab~\citep{Chen2016a,Chen2017}, where large receptive fields are achieved with hand-tuned dilation. For this experiment, we used the latest version of DeepLab v3+~\citep{Chen2018a} that incorporates the following improvements for semantic segmentation: (a) output stride of 16, (b) atrous spatial pyramid pooling (ASPP) layer, (c) global image-pool\-ing features, and (d) an output decoder layer. As a backbone network, we used ResNet101 that was modified for a semantic segmentation problem from the previous subsection.

\begin{figure}
\centering
\sidecaption
\includegraphics[width=\linewidth]{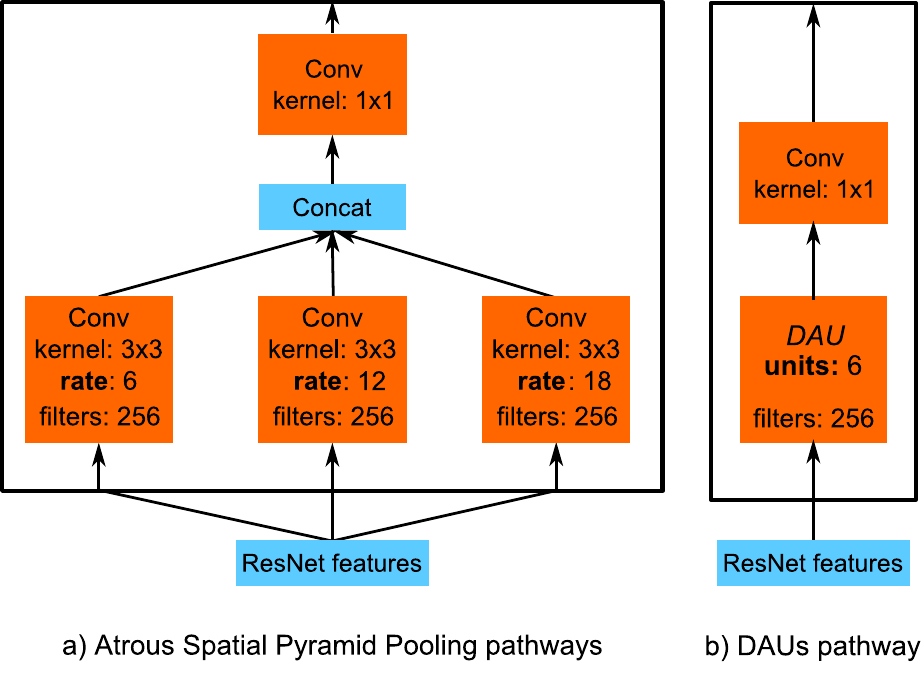}
\caption{ a) Atrous Spatial Pyramid Pooling (ASPP) block processes the input features along several parallel pathways, each containing 256 convolution filters with a fixed dilation rate -- the rates differ between the pathways. b) A single DAU pathway containing 256 DAU filters with 6 DAUs per channel  outperforms ASPP and eliminates hand-tuning of the dilation rates using less parameters. \label{fig:aspp}}
\end{figure}

The DeepLab architecture was modified to include DAUs as follows. First, the convolution filters in ResNet were replaced by DAUs in the same manner as for the DAU-ResNet from the previous subsection. Next, the atrous spatial pyramid pooling (ASPP)~\citep{Chen2016a} with three parallel convolutions, each with a different dilation rate, was replaced by a single DAU layer with six units per kernel (termed as DAU-6U) as depicted in Figure~\ref{fig:aspp}. We used more units than on other layers to provide enough coverage for larger area. Receptive field size is thus adjusted dynamically during training, provided that large enough displacement of a unit is allowed. Lastly, the output decoder was implemented with DAUs (two units per convolution filter) instead of using $3\times3$ convolutions.  

\paragraph{Dataset and Optimization.} Optimization and evaluation of DeepLab was performed on the Cityscape dataset using the same hyper-parameters as in the previous subsection. The same process of augmentation was used with the input scaling, cropping and flipping. Testing was performed on \textit{val} set and a single-scale without multi-scale testing.

\begin{figure*}
\includegraphics[width=\linewidth]{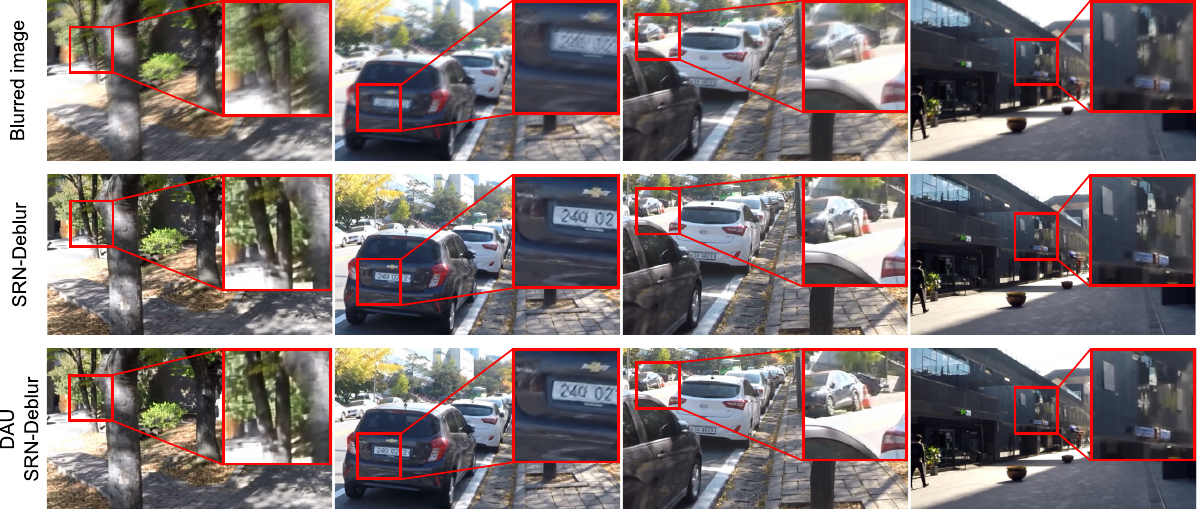}
\caption{Examples of de-blurring on GOPRO dataset with SRN-DeblurNet in the second row and DAU variant in the third row.\label{fig:deblur-examples}}
\end{figure*} 

\paragraph{Results.}
Results are reported in Table~\ref{tab:seg-results-resnet}. Results for several DeepLab versions are reported to quantify contributions of different improvements. Notice that standard ResNet101 becomes competitive with the DAU version only when image-pooling features are included. Since image-pooling features capture global information (i.e., context), this indicates that DAU convolution filters already capture most of the global information through their sparse and adjustable displacements. Results also show that a single DAU-6U layer provides a comparable performance boost to the ASPP with three hand-crafted parallel convolutions. Finally, implementing the decoder with DAUs also improves DeepLab slightly more than using a standard decoder and ASPP. In this case, the DAU version achieves a mIoU of 76.1\%, while standard DeepLab achieves a mIoU of 75.8\%.

Note that DAU-6U is a single layer with 6 units (18 parameters) while ASPP applies at least three parallel convolutions each with 9 parameters, resulting in at least 27 parameters. While ASPP was hand-crafted and would require separately testing various variations of the receptive field combinations to fine-tune its architecture to a given dataset, the DAU-6U learns them directly from the dataset, thus significantly reducing the complexity of designing high-per\-for\-mance networks.

\section{Application to Blind Image De-blurring\label{sec:deblur}}

As the last application example, we demonstrate the performance of DAUs on the task of blind image de-blurring where large receptive fields have proven to play an important role.

\subsection{A Scale-Recurrent Network with DAUs}

A scale-recurrent network by \cite{Tao2018}, termed SRN-DeblurNet, is a state-of-the-art method for blind image de-bluring task. SRN-DeblurNet employs a $43$-layer U-Net architecture in a scale-recurrent approach to perform a dense regression of each output pixel value. SRN-DeblurNet attains large receptive field sizes by down-sampling and $5\times5$ convolution kernels. The network obtains top performance on de-blurring benchmarks~\citep{Tao2018}, but at a cost of inefficient use of parameters for the spatial coverage.

We propose a DAU-SRN-DeblurNet where $5\times5$ convolutions are replaced with two displaced aggregation units per convolution filter. The replacements are made in all but four layers: we retain two de-convolution layers and the first and the last layers as classical convolutions\footnote{Current implementation of DAUs requires an even number of channels.}. This results in a much more efficient network with 4$\times$ fewer parameters than SRN-DeblurNet, and in per-filter adapted receptive field sizes. A central requirement of the convolution filters in SRN-DeblurNet is to enable modeling the identity function which allows the network to pass through pixels that are not blurred. Thus the standard deviations in DAU weights were reduced to $0.35$ to reduce their aggregation effect.  According to further evaluation by \cite{Tao2018} after the paper acceptance\footnote{https://github.com/jiangsutx/SRN-Deblur}, they removed the color ringing artifacts to further improve the performance by applying SRN-DeblurNet without LSTM to the RGB data. We used the same approach in our experiments.

\paragraph{The GOPRO Dataset.}
The GOPRO dataset~\citep{Nah2017} was used for training and testing. The dataset contains 2,103 pairs of training images and 1,111 pairs of testing images. Each pair consists of two colored images: a blurred image (input) and a sharp image (groundtruth), both in $1280\times720$ resolution.

\paragraph{Optimization.}
The training protocol of~\cite{Tao2018} was followed. We trained with a mini-batch stochastic gradient descent using the Adam solver~\citep{Kingma2015} for 2000 epochs with a batch size of 16 images. For SRN-Deblur\-Net, we used the best hyper-parameters provided by~\cite{Tao2018}, and a learning rate of $10^{-4}$ with a polynomial decay using a power factor of 0.3. The learning rate was increased for DAU-SRN-Deblur\-Net to $5\cdot10^{-4}$ to compensate for the smaller weights in DAUs due to normalized Gaussian blurring. Furthermore, since the unit displacement values are several orders of magnitude larger than the weights, we also increased the learning rate for displacement values $\mu$ to $10^{-3}$ and applied a linear decay, i.e., a polynomial decay with a power factor of 1.0. The trainable variables were initialized with the \cite{Glorot2010} method. Displacement values of DAUs were initialized randomly with a zero-mean normal distribution with $\sigma=0.5$. Data augmentation was not used in training but images were randomly cropped into $256\times256$ patches to fit them into the memory. This followed the learning protocol of~\cite{Tao2018}.

\begin{table}
\centering
\caption{Results on the GOPRO dataset using the reference and DAU-based SRN-DeblurNet architecture with reported peak-signal-to-noise-ratio (PSNR) and the number of trainable parameters (in million). \label{tab:deblur-results}}
\begin{tabularx}{\columnwidth}{Xccc}
\toprule
 \textit{Network architecture} &  \makecell[c]{PSNR (dB)} & \makecell[c]{Number of\\params.} & \makecell[c]{Number of\\units} \\ 
\midrule
SRN-DeblurNet & \textbf{30.07}  &  6.878 M & 6.878 M  \\
DAU-SRN-DeblurNet & 30.02 & \textbf{1.781} M & \textbf{0.708} M \\
\bottomrule
\end{tabularx}

\end{table}

\paragraph{Results.}
Results are reported in Table~\ref{tab:deblur-results}. Both methods achieved a peak-signal-to-noise-ratio (PSNR) of slightly above 30 dB. Note that SRN-DeblurNet required 6.8 million parameters, while DAU-SRN-DeblurNet required only $25$\% of parameters (1.7 million) for the same performance. The difference is even more substantial when considering the number of units required for spatial coverage (see Table~\ref{tab:deblur-results}) -- in this case DAU-SRN-DeblurNet requires only $10\%$ of units compared to SRN-DeblurNet. Examples of de-blurred images with both methods are shown in Figure~\ref{fig:deblur-examples}. 

We have also observed that a larger aggregation perimeter (i.e., larger standard deviation of DAUs) did not significantly affect the performance. DAUs with $\sigma=0.5$ achieved PSNR of 29.84 dB, while DAUs with $\sigma=0.35$ resulted in PSNR of 30.02 dB. Considering that the increased aggregation perimeter introduces significant feature blurring, a larger performance difference might be expected. The small difference points to an effective and robust DAU structure that is able to compensate for the added blurring effect, which is particularly important in the deblurring task.

\section{Discussion and Conclusion~\label{sec:conclusion}}

We proposed  DAU convolution filters to replace fixed grid-based filters in classical convolutional networks. 
The DAUs modify only the convolutional layer in standard ConvNets, can be seamlessly  integrated into existing architectures, and afford several advantages.
In addition to the filter unit weights, they allow learning the receptive field size. Since the number of parameters is decoupled from the receptive field size, they efficiently allocate the free parameters, resulting in compact models and efficient learning. In addition, DAUs eliminate the need for hand-crafted convolution filter patterns (e.g., dilated convolutions) and allow automatic adaptation for a broad spectrum of computer vision tasks. 

The parameter reduction capability was demonstrated on classification, semantic segmentation and blind image de-blurring. In particular, experiments with the AlexNet~\citep{Krizhevsky2012} architecture on a classification task have shown that DAUs achieved a similar performance to the standard network with only 30\% of the parameters. Similar improvement has been demonstrated on a state-of-the-art de-blurring method, SRN-DeblurNet~\citep{Tao2018}, where the same performance has been achieved with only $25\%$ of the parameters. With only three free parameters per unit, this shows that networks can potentially allocate an order of magnitude fewer units for providing sufficient spatial coverage and that existing deep learning methods are inefficiently using their parameters. 

The experiments on semantic segmentation have further shown that DAUs trained for one task enable a straightforward adaptation to another task by using the same architectural model and only learning new parameters for the new task. We have demonstrated this by adapting a DAU-based residual network~\citep{He2015a} with the architecture for the classification to the semantic segmentation. The experiment on Cityscape dataset has shown improvement in the performance of the DAU-ResNet model by 4\% compared to standard ResNet without significant modifications to the network. This was achieved while using 15\% fewer parameters. The classic ResNet has become competitive to DAUs only after global image-pooling features have been added. This means that DAUs already capture the contextual information through position adaptation, which has to be added manually by architectural change in the standard network. 
 
Experiments show that DAUs can completely replace atrous spatial pyramid pooling (ASPP) in DeepLab \citep{Cheng2014}. By adjusting displacement, DAUs were able to selectively focus on spatial areas of sub-features that are important for specific tasks. This was demonstrated on semantic segmentation where DeepLab with only a single extra DAU layer was able to fully replace several parallel convolutions in ASPP that use different dilation factors. This was achieved at the same or slightly better performance while using 15\% fewer parameters. More importantly, DAUs removed the need for hand-tuning the dilation factors in Deep\-Lab. Thus, they enable learning without repeating extensive experiments to hand-tune dilation for a new domain.

DAUs seamlessly integrate into existing state-of-the-art architectures with plug-and-play capability by simply replacing the standard convolution layers. We have published CUDA implementations for Caffe and TensorFlow frameworks and plan to release all the DAU versions of state-of-the-art architectures reported in this work, making all results in this work fully reproducible. 
 
An active area of exploration in the deep learning community is development of mathematical tools for formal analysis of ConvNet properties.
Such analysis is very difficult with the classical ConvNet formulation with discrete convolution filters that can take arbitrary values, and simplifications of the model have to be made. A highly interesting analysis is the work of~\cite{Bruna2013} who 
treat ConvNets from a spectral perspective. Note that DAUs also provide a new formal view of the ConvNet pipeline. In their simplest variant with a single unit per convolution filter, DAU-ConvNets can be considered as a sequence of feature low-pass filtering (blurring) and spatial shifting with intermediate nonlinearities. Our results show that even this simplest formulation achieves comparable performance to the classical ConvNets, but is more tractable. We expect that this mathematically simpler formulation will open new venues for further theoretical analysis of deep models.

\section{Acknowledgements}
The authors would like to thank Hector Basevi for his valuable comments and suggecstion on improving the paper. This work was supported in part by the following research projects and programs: project GOSTOP C3330-16-529000, DIVID J2-9433 and ViAMaRo L2-6765, program P2-0214 financed by Slovenian Research Agency ARRS, and MURI project financed by MoD/Dstl and EPSRC through EP/N019415/1 grant. We thank Vitjan Zavrtanik for his contribution in porting the DAUs to the TensorFlow framework.

{\small
\bibliographystyle{spbasic}
\bibliography{library}
}

\end{document}